\documentclass{article}

\usepackage[final]{neurips_2025}

\usepackage{hyperref}
\usepackage{adjustbox}
\usepackage{graphicx}
\usepackage{caption}
\usepackage{subcaption}
\usepackage{printlen}
\usepackage{float}
\usepackage{wrapfig}
\usepackage{multirow}
\usepackage{booktabs}
\usepackage{pifont}
\usepackage{makecell}
\usepackage{nicefrac}
\usepackage{verbatim}
\usepackage{amsmath}
\usepackage{algorithm}
\usepackage{algpseudocode}
\usepackage{amsfonts}
\usepackage{microtype}
\usepackage{pgfplots}
\usepackage{tabularx}
\usepackage{ifthen}
\usepackage{soul}
\usepackage{tikz}
\usetikzlibrary{angles,quotes,3d,math,arrows.meta,calc,positioning,fit,backgrounds,decorations.pathreplacing,calligraphy,shapes,shapes.multipart}
\usepackage[table]{xcolor}
\usepackage[capitalize]{cleveref}
\usepackage{enumitem}

\newcommand{\videoframe}{\mathbf{x}}
\newcommand{\video}{\mathbf{X}}
\newcommand{\motion}{\mathbf{m}}
\newcommand{\motionsequence}{\mathbf{M}}
\newcommand{\motionextractor}{\mathcal{M}_\theta}

\newcommand{\timesteps}{\mathbf{t}}
\newcommand{\timesource}{{t}}
\newcommand{\deltatime}{{\Delta_t}}
\newcommand{\timedestination}{{t+\deltatime}}

\newcommand{\framegenerator}{\mathcal{F}_\psi}

\makeatletter
\newcommand{\xleftrightarrow}[2][]{\ext@arrow 3359\leftrightarrowfill@{#1}{#2}}
\newcommand{\xdashrightarrow}[2][]{\ext@arrow 0359\rightarrowfill@@{#1}{#2}}
\newcommand{\xdashleftarrow}[2][]{\ext@arrow 3095\leftarrowfill@@{#1}{#2}}
\newcommand{\xdashleftrightarrow}[2][]{\ext@arrow 3359\leftrightarrowfill@@{#1}{#2}}
\def\rightarrowfill@@{\arrowfill@@\relax\relbar\rightarrow}
\def\leftarrowfill@@{\arrowfill@@\leftarrow\relbar\relax}
\def\leftrightarrowfill@@{\arrowfill@@\leftarrow\relbar\rightarrow}
\def\arrowfill@@#1#2#3#4{%
  $\m@th\thickmuskip0mu\medmuskip\thickmuskip\thinmuskip\thickmuskip
   \relax#4#1
   \xleaders\hbox{$#4#2$}\hfill
   #3$%
}
\makeatother

\newcolumntype{H}{>{\setbox0=\hbox\bgroup}c<{\egroup}@{}}

\newcommand{\tikzstylenodedistance}{4mm}
\newcommand{\tikzstyleinnersep}{2mm}
\newcommand{\tikzstyleminimumheight}{8.75mm}
\newcommand{\tikzstyleminimumwidth}{12mm}

\tikzset{
    node distance=\tikzstylenodedistance,
    text centered,
    anchor=center,
}
\tikzset{
    standard node/.style n args={1}{%
        rectangle,
        rounded corners=0.1cm,
        fill=our#1,
        draw=our#1border,
        line width=0.04cm,
        minimum height=\tikzstyleminimumheight,
        minimum width=\tikzstyleminimumwidth,
        inner sep=\tikzstyleinnersep,
        text centered,
        anchor=center,
        align=center,
    }
}
\tikzset{
    standard node module/.style n args={0}{%
        rectangle,
        rounded corners=0.1cm,
        fill=ourturquoise,
        draw=ourturquoiseborder,
        line width=0.04cm,
        minimum height=\tikzstyleminimumheight, %
        minimum width=12mm, %
        inner xsep=\tikzstyleinnersep,
        inner ysep=1mm,
        text centered,
        anchor=center,
        align=center,
    }
}
\tikzset{
    standard node image/.style n args={1}{%
        rectangle,
        fill=our#1,
        draw=our#1border,
        line width=0.04cm,
        minimum height=\tikzstyleminimumheight,
        minimum width=\tikzstyleminimumwidth,
        inner sep=0,
        text centered,
        anchor=center,
        align=center,
    }
}
\tikzset{
    standard node circle/.style n args={1}{%
        fill=our#1,
        draw=our#1border,
        circle,
        inner sep=0.1cm,
        minimum height=0,
        minimum width=0,
    }
}
\tikzset{
    standard node circle/.prefix style = standard node
}

\tikzset{
    standard line/.style n args={0}{%
        line width=0.04cm,
        rounded corners=0.1cm,
    }
}
\tikzset{
    standard arrow/.style n args={0}{%
        -latex,
    }
}
\tikzset{
    standard arrow/.prefix style = standard line
}

\tikzset{
    simple node image/.style n args={0}{%
        rectangle,
        inner sep=0,
        text centered,
        anchor=center,
        align=center,
        node distance=0mm
    }
}

\usepackage[utf8]{inputenc} %
\usepackage[T1]{fontenc}    %
\usepackage{hyperref}       %
\usepackage{url}            %
\usepackage{booktabs}       %
\usepackage{amsfonts}       %
\usepackage{nicefrac}       %
\usepackage{microtype}      %
\usepackage{xcolor}         %

\title{\textit{DisMo}: Disentangled Motion Representations\\for Open-World Motion Transfer}

\author{%
\textbf{Thomas Ressler-Antal} \quad \textbf{Frank Fundel}\footnote[1]{test} \quad \textbf{Malek Ben Alaya}$^{*}$ \\ \textbf{Stefan Andreas Baumann} \quad \textbf{Felix Krause} \quad \textbf{Ming Gui} \quad \textbf{Björn Ommer} \\\\
CompVis @ LMU Munich, Munich Center for Machine Learning (MCML)
}

\begin{document}

\maketitle

\def\thefootnote{*}\footnotetext{Equal contribution.}\def\thefootnote{\arabic{footnote}}

\begin{figure}[ht]
  \centering
  \includegraphics[width=1\linewidth]{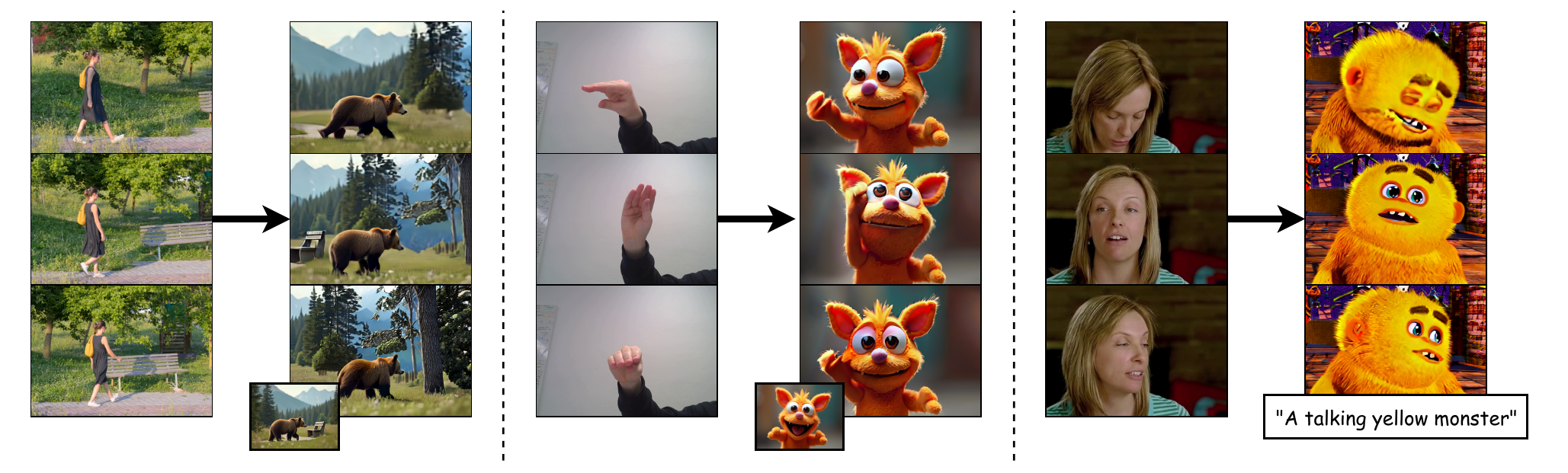}
  \caption{\textbf{Motion transfer examples enabled by our abstract motion representations.} We extract abstract motion representations from driving videos and transfer them onto new content, represented either by source images \textbf{(Left)} and \textbf{(Middle)}, or text prompts \textbf{(Right)}. }
\end{figure}
\begin{abstract}
Recent advances in text-to-video (T2V) and image-to-video (I2V) models, have enabled the creation of visually compelling and dynamic videos from simple textual descriptions or initial frames. However, these models often fail to provide an explicit representation of motion separate from content, limiting their applicability for content creators. To address this gap, we propose \textit{DisMo}, a novel paradigm for learning abstract motion representations directly from raw video data via an image-space reconstruction objective. Our representation is generic and independent of static information such as appearance, object identity, or pose. This enables open-world motion transfer, allowing motion to be transferred across semantically unrelated entities without requiring object correspondences, even between vastly different categories. Unlike prior methods, which trade off motion fidelity and prompt adherence, are overfitting to source structure or drifting from the described action, our approach disentangles motion semantics from appearance, enabling accurate transfer and faithful conditioning. Furthermore, our motion representation can be combined with any existing video generator via lightweight adapters, allowing us to effortlessly benefit from future advancements in video models. We demonstrate the effectiveness of our method through a diverse set of motion transfer tasks. Finally, we show that the learned representations are well-suited for downstream motion understanding tasks, consistently outperforming state-of-the-art video representation models such as V-JEPA in zero-shot action classification on benchmarks including Something-Something v2 and Jester. Project page: \url{https://compvis.github.io/DisMo}
\end{abstract}
    
\section{Introduction}\label{sec:intro}

A striking feature of human visual intelligence is our innate ability to understand motion and to reason about it independently from static attributes such as appearance, position, and viewpoint, generalizing seamlessly across different instances within an object category and even beyond to entirely different objects and scenarios  \citep{johansson1973visual}. Already at a young age, we are able to generalize motion despite stark deviations in environment and content to a different scene allowing already little children to imagine the plausible movement of fictitious characters. This capacity highlights a fundamental cognitive ability to represent and reason about motion as an abstract, invariant phenomenon.

Recent breakthroughs in generative video synthesis have enabled models to produce visually compelling and dynamically realistic videos from minimal inputs such as textual descriptions (text-to-video, T2V) or single initial frames (image-to-video, I2V) \citep{blattmann2023stable, brooks2024sora, HaCohen2024LTXVideo, kong2024hunyuanvideo, veo2}. Despite the remarkable rendering quality achieved by these models, they lack explicit representations of underlying kinematics, inherently limiting their ability to precisely control or manipulate the motion of generated scenes.

Existing T2V and I2V approaches implicitly represent motion by modeling videos as sequences of static frames, which aligns poorly with human perception that inherently treats motion as a distinct and explicitly perceived entity rather than merely a tomography across time \citep{kandel:neural}. Furthermore, precise textual descriptions of complex or abstract motion remain challenging, underscoring the need for novel representation approaches that go beyond verbal or purely frame-based encodings.

Explicit methods for representing dynamic scene content have traditionally relied on dense, pixel-level motion descriptors such as optical flow \citep{zhao2024motiondirector, chefer2025videojam, liang2024movideo}. While optical flow effectively captures detailed per-pixel changes between frames, it remains strongly coupled to specific instance structures, locations, and viewpoints, thus lacking the invariance needed for robust motion generalization and transfer across diverse scenarios. Similarly, sparse representations such as trackers that follow key points through time are somewhat less dependent on appearance but still fundamentally require structural consistency across frames \citep{karaev2024cotracker}. Consequently, neither dense nor sparse methods readily support motion transfer between significantly different instances, object categories, and viewpoints.

Inspired by recent successes in text-to-image (T2I) synthesis, where explicit structural control mechanisms have greatly enhanced user-driven generation \citep{rombach2022ldm, zhang2023adding, stracke2024ctrloralter}, we argue that generative video models similarly require novel mechanisms for explicitly representing and controlling motion separate of appearance and spatial configurations. Such explicit and abstract representations would significantly enhance user control over dynamically generated content.

In this work, we introduce a method for learning abstract motion representations that are invariant to static appearance and structure-based information, directly from raw video data using only a standard image reconstruction loss. Remarkably, our representation emerges entirely through end-to-end training regime without the need for specialized regularization, contrastive losses, or complex training setups. We leverage a simple yet effective flow-matching-based reconstruction objective, commonly used in visual generation methods, to obtain a representation that is largely invariant to factors typically encoded in the reconstruction loss, such as position, identity, category, and pose. The result is a motion representation robust enough to perform challenging and creative open-world motion transfers, enabling scenarios such as transferring the movement of a human to that of an ape, an imaginary cartoon character, or even completely unconventional transformations. This capability is achieved through an information bottleneck further by strong augmentations that makes it impossible to reconstruct the target using visual and appearance-based information alone, forcing the model to analyze changes between frames and encode them into an abstract embedding.

We evaluate these embeddings via zero-shot probing across multiple action classification datasets and ensure their representative performance. Additionally, we design these motion representations to be invariant to the choice of renderer, allowing seamless integration with state-of-the-art video diffusion models through lightweight adapters~\citep{stracke2024ctrloralter}, which reduces compute as expensive video pipelines do not need to be retrained. This orthogonal approach transforms existing generative video models from merely powerful renderers into highly controllable generative systems.

Our main contributions are as follows:
\begin{itemize}[leftmargin=15pt]
    \item We construct a powerful representation learning paradigm powered only by an image-space reconstruction objective and show how this leads to an abstract motion representation space that is trained for invariance to static content information such as appearance or structure.
    \item We leverage this invariance and condition a pre-trained generative video model on our learned motion representations, which, despite training in a matched setting is capable of zero-shot motion transfer.
    \item We show how this setup achieves state-of-the-art performance on open-world motion transfer with a high degree of transferability, especially in cross-category and cross-viewpoint settings.
    \item We show how our motion embeddings can be utilized in downstream action classification settings.
\end{itemize}

\section{Related Work}

\paragraph{Video Synthesis}
Generative image and video models have experienced rapid advances in recent years, largely driven by diffusion-based training objectives~\citep{ho2020denoising,song2020score,lipman2022flow}. Originally developed for the image domain, these powerful frameworks have been extended to the video domain, demonstrating impressive generative capabilities and realism. \citet{brooks2024sora} demonstrated that large-scale video synthesis enables the realistic generation of complex spatiotemporal dynamics from simple text and image prompts. Subsequently, several state-of-the-art text-to-video (T2V) and image-to-video (I2V) diffusion models have further significantly advanced video synthesis capabilities \citep{blattmann2023stable,yang2024cogvideox,HaCohen2024LTXVideo,kong2024hunyuanvideo,veo2,polyak2025moviegencastmedia} and made such models widely accessible.

Despite their success, current T2V/I2V methods primarily focus on synthesizing visually appealing videos from textual and image prompts but do not explicitly model the underlying motion separately from appearance. As a result, they inherently lack the capacity for precise or intuitive control over the generated motion, limiting their applicability in scenarios requiring detailed, user-specified dynamics.

\paragraph{Motion Control in Video Generation}
Enabling detailed motion control in generative video models remains a challenging and active research area. Recent methods typically achieve motion control by providing explicit low-level spatial or temporal constraints, such as bounding boxes \citep{jain2024peekaboo,wang2024boximator,wu2025motionbooth}, trajectories \citep{wang2024motionctrl,wu2024draganything,yin2023dragnuwa}, or structural maps like optical flow and depth sequences \citep{zhang2023controlvideo,esser2023structure}. Although effective within certain scenarios, these conditioning signals inherently tie motion control to precise spatial locations and/or structures in the source video, limiting their invariance and transferability. Consequently, these approaches struggle with scenarios involving substantial variations in viewpoint, appearance, or semantic category.

Parametric approaches offer another perspective by learning domain-specific motion representations, frequently used in specialized areas such as facial reenactment and talking-head generation \citep{siarohin2019first,siarohin2021motion,wang2021one,zhao2022thin,wang2022latent,hong2022depth,yang2022face2face,gao2023high}. These methods typically rely on pre-defined or learned keypoint-based \textit{category-specific} representations (either 2D or 3D) to encode motion explicitly. For instance, First-Order Motion Models (FOMM) \citep{siarohin2019first}, Motion Representations for Articulated Animation (MRAA) \citep{siarohin2021motion}, and Latent Image Animator (LIA) \citep{wang2022latent} have demonstrated impressive performance. However, due to their specialized nature and dependence on predefined structural cues, parametric models have to be specifically developed for the target domain/category and typically generalize poorly beyond their narrowly defined application, not allowing for advanced applications such as cross-category motion transfer.

In contrast to both explicit low-level conditioning and domain-specific parametric models, our approach aims to learn a highly abstract, invariant motion representation. By explicitly modeling motion independently of appearance, pose, viewpoint, and semantic category, we substantially enhance the flexibility and transferability required for diverse motion control scenarios.

\begin{figure*}[t]
    \vspace{1em}
    \centering
    \includegraphics[width=\linewidth]{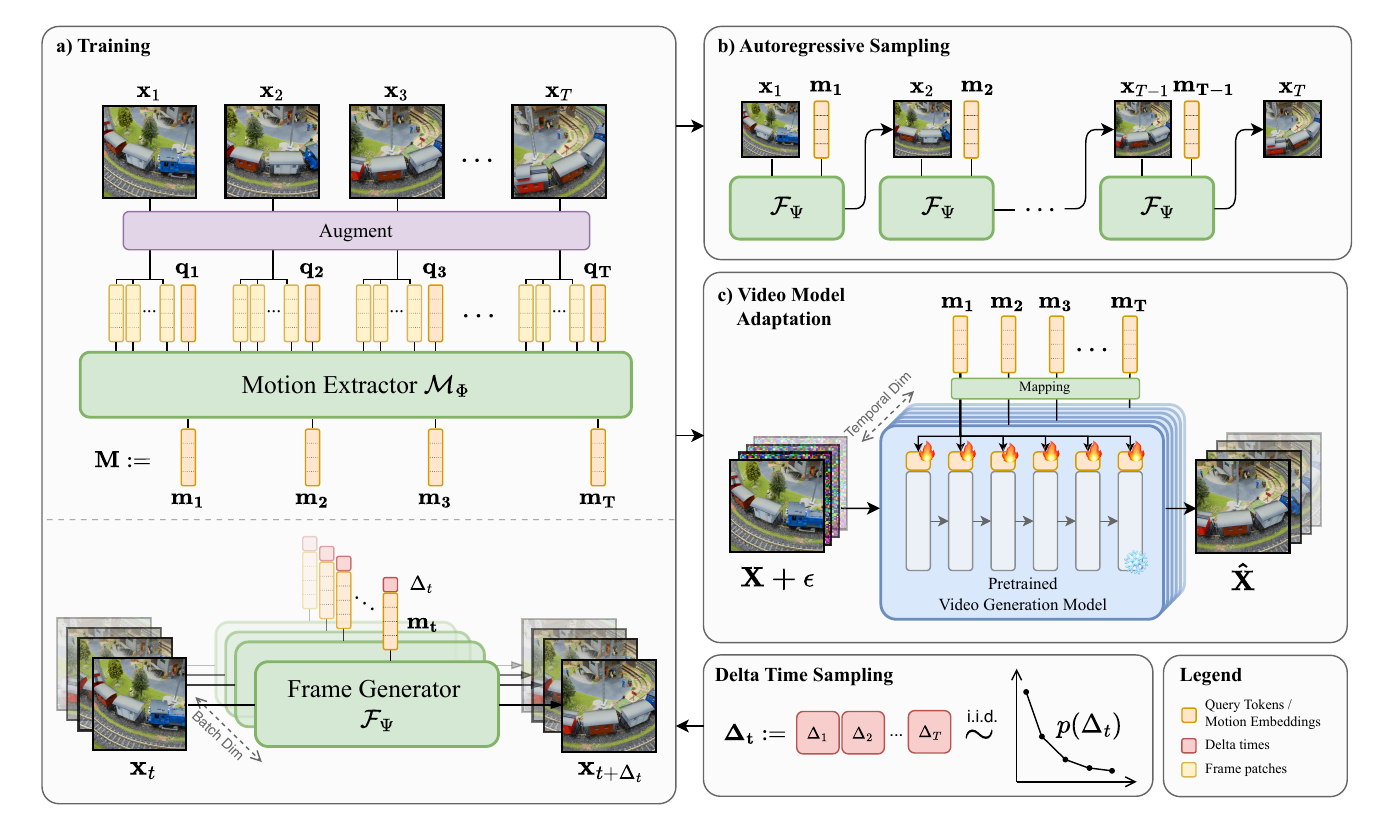}
    \caption{\textbf{Method Overview.} \textbf{(a)} During training, our motion extractor $\motionextractor$ receives augmented frames from a video $\video$, along with additional motion query tokens $\mathbf{Q}$. These are then individually passed to the frame generator $\framegenerator$, alongside the corresponding source frame \(\videoframe_t\), from which it learns to reconstruct a frame at a future timestep $\timedestination$. 
    \textbf{(b)} To transfer a motion sequence $\motionsequence$ onto another target image, we can directly utilize the trained Frame Generator $\mathcal{F}_{\psi}$ autoregressively as a low-cost option.
    \textbf{(c)} For high-quality motion transfer, we adapt pre-trained off-the-shelf video generation models. A motion sequence is first embedded using a mapping network, before being introduced to the frozen video model. The processed motion sequence is arranged such that each token at timestep \(t\) in the pre-trained backbone receives conditioning only from the temporally corresponding motion embedding $\motion_t$.}
    \label{fig:method}
\end{figure*}

\section{Method}

Recent approaches in video modeling commonly utilize generative diffusion or flow-based objectives applied either directly on video frames~\citep{ho2022imagenvideohighdefinition} or on latent representations extracted from pretrained autoencoders \citep{blattmann2023stable,brooks2024sora,HaCohen2024LTXVideo,kong2024hunyuanvideo,veo2}. While these have been successful for high-fidelity synthesis tasks, our goal differs fundamentally: instead of directly generating videos, we aim to learn abstract, motion representations that are invariant to the appearance of the scene. Specifically, we define motion as the abstract temporal dynamics that describe \textit{how} scenes and the objects therein evolve, independent from their actual appearance, style, spatial configuration, or even semantic category. The representations can subsequently serve as a powerful, generalizable conditioning signal for controlling motion in downstream video generation or discriminative tasks.

\subsection{Preliminaries}
\paragraph{Flow Matching} Flow matching~\citep{lipman2022flow} is a generative modeling framework that learns distributions via a continuous-time flow field that smoothly transforms a simple prior distribution--typically a Gaussian--into a complex target data distribution. Given data samples $\mathbf{z} \sim p_\mathrm{data}(\mathbf{z})$
, they learn a vector field $\mathbf{v}_\psi(\mathbf{z}_\tau, \tau)$ with learnable parameters $\psi$ describing the flow of the distribution at intermediate times $\tau \in [0, 1]$. This vector field is learned by minimizing the MSE loss
\begin{equation}\label{eq:flow_matching}
    \mathcal{L}(\psi) = \mathbb{E}_{\tau, \mathbf{z}_0, \mathbf{z}_1} \|\mathbf{v}_\psi(\mathbf{z}_\tau, \tau) - \mathbf{u}(\mathbf{z}_\tau, \tau) \|_2^2,
\end{equation}
where $\mathbf{z}_\tau = \tau \mathbf{z}_1 + (1 - \tau)\mathbf{z}_0, \mathbf{z}_1 \sim p_\mathrm{data}$, $\mathbf{u}(\mathbf{z}_\tau, \tau) = \mathbf{z}_1 - \mathbf{z}_0$ is the target flow field, and $\mathbf{z}_0$ is drawn from the prior distribution, which is typically chosen to be Gaussian $\mathbf{z}_0 \sim \mathcal{N}(0, \mathbf{I})$.
Samples from this learned distribution can be generated by drawing from the prior and solving the learned ODE.

\subsection{Learning Transferable Motion Representations} \label{sec:video_model_adaption}
To learn abstract motion representations, we introduce \textit{DisMo}, a method consisting of two primary components learned jointly: a motion extractor $\motionextractor$ and a motion-conditioned frame generator $\framegenerator$. The motion extractor computes meaningful motion embeddings from raw input videos, while the frame generator leverages these embeddings alongside source frames to reconstruct target frames. The overall framework is trained end-to-end using a simple yet powerful generative image-space reconstruction objective, modeling the distribution of frames conditioned on a source frame and learned motion embedding. The learned motion embedding inherently encodes temporal dynamics disentangled from frame contents and appearance. An overview of the architecture and training paradigm is depicted in \cref{fig:method}.

\paragraph{Motion Extraction.}
Consider an input video $\video = \{\videoframe_{t}\}_{t=1}^{T}$ consisting of $T$ frames $\videoframe_{t}$. From this, our goal is to extract a sequence of motion embeddings $\motionsequence = \{\motion_{t}\}_{t=1}^{T}$. We formally define the motion extractor as $\motionextractor: \video\rightarrow \motionsequence$ with parameters $\mathbf{\theta}$. This motion extractor jointly processes the given frames and extracts a sequence of motion embeddings for all timesteps $\timesteps = \{1, \ldots, T\}$. Each motion embedding encodes local temporal dynamics while ignoring static visual information.

\paragraph{Dual-Stream Frame Generation.}
The frame generator $\framegenerator$, parametrized by $\psi$, reconstructs target frames conditioned on a motion embedding and a source frame, and is responsible for providing the learning signal for the motion extractor. Formally, the generator learns the conditional distribution
\begin{equation}
    \framegenerator: \videoframe_\timedestination \sim p(\videoframe_\timedestination \mid \videoframe_\timesource, \motion_\timesource, \deltatime)
\end{equation}
describing the future destination frame $\videoframe_\timedestination$ at timestep $\timedestination$. As conditioning, the source frame $\videoframe_\timesource$, the motion embedding at the source time $\motion_\timesource$ extracted using our motion extractor $\motionextractor$ and the distance $\deltatime$ to the target frame are given. $\deltatime$ can either be a constant or can be sampled independently for each prediction. Effectively, this provides the frame generator with two streams of available information: content information captured by the source frame $\videoframe_\timesource$, and information that is embedded into the motion representation $\motion_\timesource$. Due to the information bottleneck induced on the motion embedding via the limited embedding size, this representation is encouraged to store only high-level, invariant temporal information. Essentially, this represents the residual knowledge not encoded by $\videoframe_\timesource$ that supports the frame generator in predicting the target frame $\videoframe_\timedestination$.

We use this frame generator $\framegenerator$ to jointly learn both it and the motion extractor $\motionextractor$ end-to-end by (implicitly) maximizing the likelihood~\citep{kingma2023understanding} of the reconstructed destination frames $\videoframe_\timedestination$, conditioned on respective source frames and their corresponding motion embeddings:
\begin{equation}
    \mathbb{E}_{\video \sim p_\mathrm{data}}[\log(p_\psi(\videoframe_\timedestination \mid \videoframe_\timesource, \motionextractor(\video, \timesource))].
\end{equation}
We practically implement this via a flow matching~\citep{lipman2022flow} objective (cf. \cref{eq:flow_matching})
\begin{equation}
    \mathcal{L}(\theta,\psi) = \mathbb{E}_{\video \sim p_\mathrm{data}, \tau, \mathbf{z}_0 \sim \mathcal{N}(0, \mathbf{I}), (\timesource, \timedestination)}\\
    \|\mathbf{v}_\psi(\mathbf{z}_\tau, \videoframe_\timesource, \motionextractor(\video, \timesource), \tau) - \mathbf{u}(\mathbf{z}_\tau, \tau)\|_2^2,
\end{equation}
where $\mathbf{z}_\tau = \tau \videoframe_\timedestination + (1 - \tau)\mathbf{z}_0$, and $\mathbf{v}_\psi$ is the vector field learned by the frame generator $\framegenerator$.

\paragraph{Augmentation Pipeline.}
To ensure that motion embeddings remain invariant to appearance, we additionally employ an augmentation pipeline inspired by self-supervised learning methods \citep{he2022masked,tong2022videomae,chen2020simple}. Appearance augmentations include random photometric transformations (e.g., hue, contrast, brightness adjustments) and geometric transformations (e.g., cropping, rotation, translation, shearing, aspect ratio changes), applied uniformly across all frames to avoid confusing these changes with actual motion. By applying these augmentations during training, we further encourage the learned motion embeddings to capture high-level, transferable dynamics rather than video-specific details.

\subsection{Motion-Guided Video Generation} \label{subsec:motion_guidance}
Once the motion extractor and frame generator are trained, they can be directly used for autoregressive motion transfer. Given a source video, we extract a sequence of motion embeddings and, starting from a target initial frame, we condition the frame generator on these embeddings at each time step to predict the next frame in an autoregressive manner (see \cref{fig:method}b). While effective, this setup is limited by the temporal scope captured in a single motion embedding and the capacity of the frame generator. To address these limitations and enable longer, more coherent video generation, we further adapt off-the-shelf video generation model by conditioning them on sequences of motion embeddings. This approach allows us to leverage the strong generative priors of large video models while guiding their synthesis with our learned motion representation. 

To adapt a pre-trained video backbone, we utilize LoRAdapters~\citep{stracke2024ctrloralter}, adaptive low-rank adapters~\citep[LoRAs,][]{hu2022lora} that allow introducing additional conditioning into a frozen backbone. We extend a pre-trained spatio-temporal video model with conditional LoRAs that receive a sequence of motion embeddings during the generation process as input. Importantly, the conditioning is designed such that each spatiotemporal token gets influenced only by the motion embedding that aligns with its temporal position. This makes the approach efficient, yet effective. This fine-tuning regime is illustrated in \cref{fig:method}c. Our generated motion embeddings are generally applicable to a wide range of off-the-shelf video models. We demonstrates that DisMo can be adapted to various video models in the Appendix (see \cref{tab:video_quality}), while benefitting from their advantages.

\section{Experiments}
We evaluate our learning paradigm and show the effectiveness of the learned motion representations for motion transfer tasks. We further show that the motion embeddings are disentangled from appearance and structure-specific information, preventing unwanted information leakage during transfer tasks. Finally, we demonstrate that our motion representations encode higher-level temporal dynamics that is useful in zero-shot action classification settings.

\paragraph{Implementation Details.} We implement our motion extractor as a 3D vision transformer~\citep[][$86$M params]{dosovitskiy2021vit} with an initial frame embedding stage consisting of a DINOv2-B~\citep{oquab_dinov2_2023} model, and adapt our frame generator from a pretrained diffusion transformer~\citep[][$675$M params]{peebles2023scalable}. We train the two components using AdamW~\citep{loshchilov2019decoupledweightdecayregularization} for $530$k steps at a batch size of $32$ on open-world videos from K-710~\citep{li2022uniformerv2spatiotemporallearningarming}, SSv2~\citep{goyal2017something}, Moments in Time~\citep{monfort2019momentstimedatasetmillion} and OpenVid-1m~\citep{nan2025openvidm} to learn general-purpose motion representations. Unless otherwise specified, we use LTX-Video-2B~\citep{HaCohen2024LTXVideo} as our video generation backbone that we adapted following the procedure described in \cref{sec:video_model_adaption}. Evaluations with other video model backbones, additional implementation details and ablations of our main design choices and components are provided in the Appendix.

\subsection{Open-World Motion Transfer}
\begin{figure*}[htbp]
    \centering
    \includegraphics[width=0.85\textwidth]{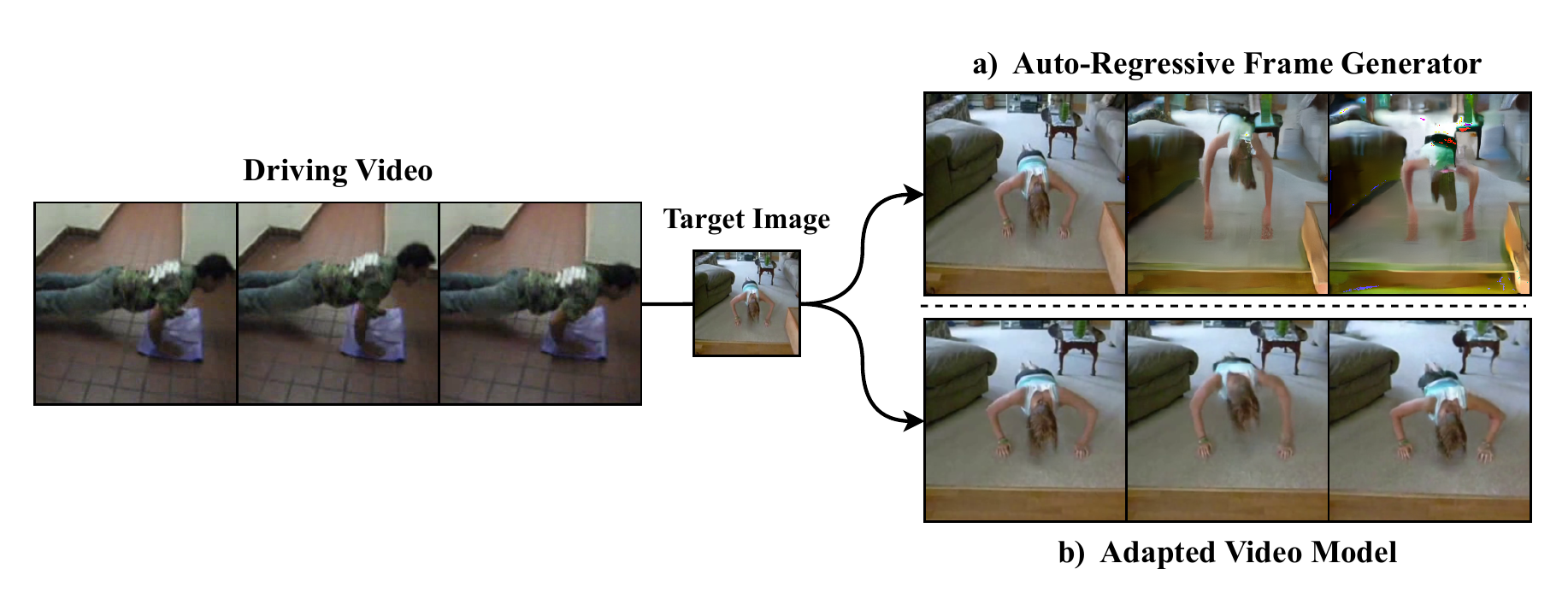}
    \caption{\textbf{Qualitative motion transfer comparison between (a) our auto-regressive frame generator  and (b) an adapted video model.} We transfer motion extracted from a driving video onto a new target image using each model. While both approaches manage to transfer high-level motion semantics in a view- and appearance-invariant manner, the adapted video model achieves higher generation fidelity.}
    \label{fig:frame_dec_vs_ltx}
\end{figure*}

We qualitatively compare our auto-regressive frame generator and the adapted high-fidelity video model in a motion transfer scenario in \cref{fig:frame_dec_vs_ltx}. While both successfully transfer higher-level motion from the source video onto the target image, the adapted video model yields a substantial improvement in generation quality. We therefore rely solely on the adapted model in the following experiments.

\begin{figure*}
  \centering
  \includegraphics[width=1\linewidth]{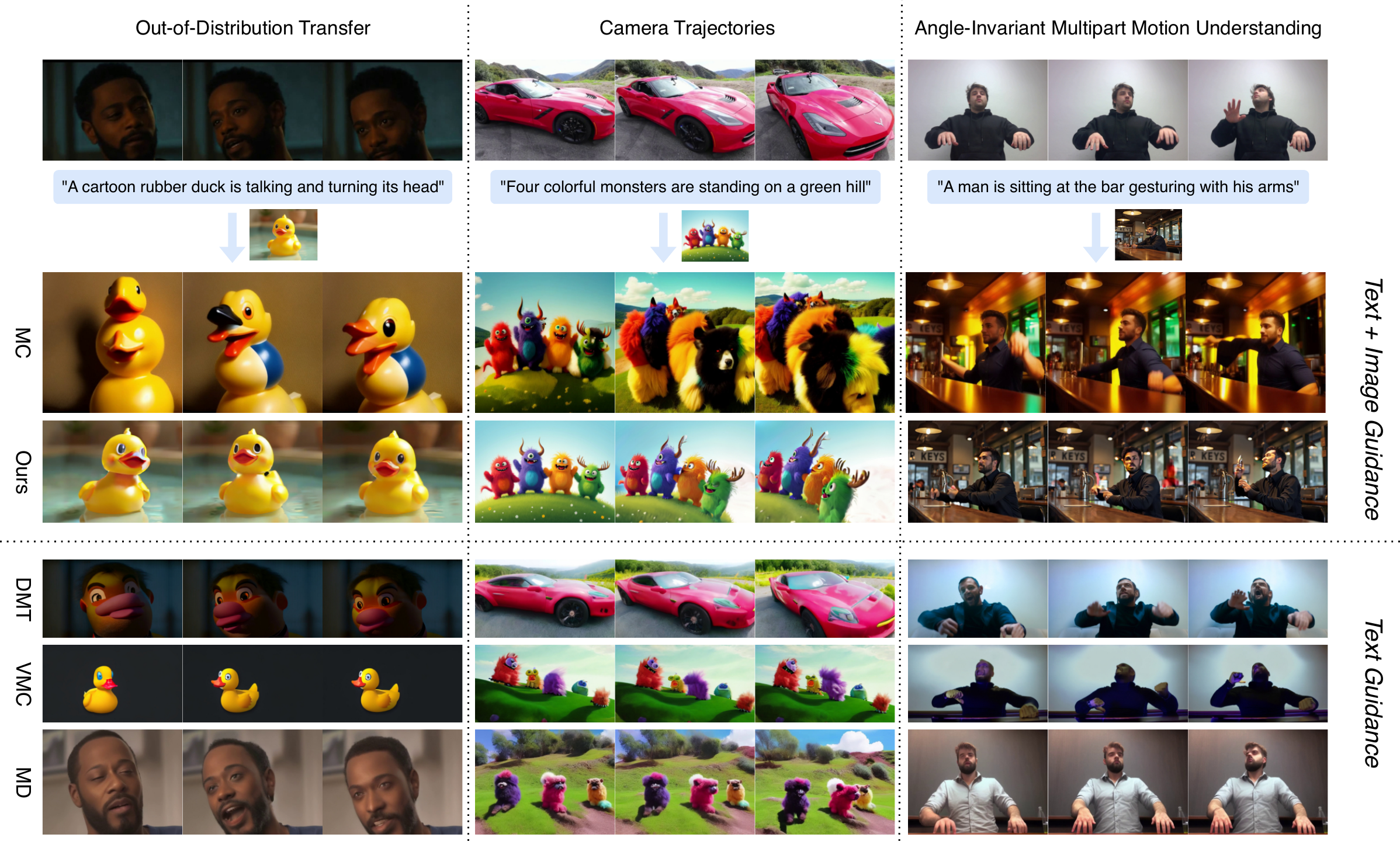}
  \caption{\textbf{Motion transfer examples in different settings.}
  We compare different motion transfer methods on three settings:
  \textbf{(Left)} Inter-category motion transfer. \textbf{(Middle)} An example showcasing camera motion. \textbf{(Right)} Composed motion transfer.}
  \label{fig:general_examples}
\end{figure*}

Specifically, we evaluate it on general and out-of-distribution motion transfer scenarios and compare it against four state-of-the-art approaches: VMC \citep{jeong2024vmc}, DMT \citep{yatim2024space}, MotionClone \citep{ling2024motionclone} and MotionDirector \citep{zhao2024motiondirector}. VMC and MotionDirector are fine-tuning-based methods that require extensive retraining per motion concept, whereas DMT and MotionClone are training-free approaches that leverage aggregated priors from a pre-trained video diffusion model for a test-time optimization objective.

Unlike many motion transfer methods, such as VMC and DMT, which rely solely on text-based appearance guidance, our method can optionally incorporate a start frame conditioning. This provides more nuanced control over the content, appearance, and structure of the target entities. However, our approach also performs effectively using only textual prompts.

To comprehensively assess method performance, we build upon prior evaluation strategies \citep{zhao2024motiondirector,jeong2024vmc,yatim2024space, park2024spectral}, assembling a diverse set of videos and text prompts primarily from the DAVIS dataset \citep{pont20172017}. Additionally, we include challenging examples from OpenVid-1M \citep{nan2025openvidm} and other sources. Our evaluation set features varied scenarios such as stationary and moving cameras, in-domain and out-of-domain motion transfers, and creatively challenging out-of-distribution cases.

We quantitatively compare methods using three established metrics: \textit{Motion Fidelity}: Measured using a tracking-based metric inspired by Chamfer distance, evaluating alignment between generated and driving video tracklets \citep{yatim2024space}. \textit{Prompt Adherence}: Evaluated by computing the cosine similarity between CLIP embeddings of generated frames and corresponding textual prompts. \textit{Temporal Consistency}: Assessed by measuring the alignment of CLIP embeddings between consecutive frames. \textit{Driving Video Similarity}: assessed by measuring alignment of CLIP embeddings of source and generated video frames. Our evaluation (see \cref{tab:general_transfer_quantitative}) shows that DisMo consistently matches or outperforms other state-of-the-art methods across all four metrics. Notably, the evaluated models exhibit inherent trade-offs: DMT and MotionDirector achieve high motion fidelity but frequently overfit on the exact structure of source videos, compromising textual adherence, while VMC and MotionClone maintain strong textual adherence at the expense of accurate motion transfer.  Meanwhile, our approach performs best on both metrics, not suffering from this trade-off. Qualitative examples in \cref{fig:general_examples} further illustrate this. Other methods either fail to transfer the object/camera motion faithfully or fail to adhere to the appearance of the prompt/start frame. \textit{DisMo}’s ability to represent motion abstractly allows greater flexibility and generalization compared to existing approaches.

\begin{table}[]
    \centering
    \caption{
    \textbf{Quantitative Motion Transfer Evaluation.}
    We compare with four other motion transfer methods with respect to motion fidelity, text adherence, temporal consistency, and similarity to the source video during our generative transfer experiments. \textit{DisMo} performs best on all four metrics, while other approaches struggle with the trade-off between motion fidelity and condition adherence.
    }
    \vspace{3mm}
    \small
    \adjustbox{max width=\columnwidth}{
        \begin{tabular}{lcccc}
            \toprule
            \small \textbf{\shortstack{Method}} &
            \small \textbf{\shortstack{Motion\\Fidelity $\uparrow$}} &
            \small \textbf{\shortstack{Prompt\\Adherence $\uparrow$}} &
            \small \textbf{\shortstack{Temporal\\Consistency $\uparrow$}} &
            \small \textbf{\shortstack{Driving Video\\Similarity $\downarrow$}} \\
            \midrule
            
            VMC*~\citep{jeong2024vmc} & 0.57 & {0.26} & \underline{0.94} & \underline{0.59} \\
            DMT$\dagger$~\citep{yatim2024space}  & \underline{0.70} & 0.24 & 0.93 & 0.66 \\
            MotionClone$\dagger$~\citep{ling2024motionclone}  & 0.63 & \textbf{0.27} & 0.91 & \underline{0.59} \\
            MotionDirector*~\citep{zhao2024motiondirector}  & \underline{0.70} & 0.16 & 0.92 & 0.82 \\
            \textit{DisMo$\ddagger$ (Ours)}  & \textbf{0.75} & \textbf{0.27} & \textbf{0.95} & \textbf{0.55} \\
            
            \bottomrule
        \end{tabular}
    }
    {\scriptsize {{*Per-sample} finetuning $\quad \dagger$Inference-time optimization $\quad \ddagger$General pre-trained}}
    \label{tab:general_transfer_quantitative}
\end{table}

Furthermore, perceptual aspects such as motion transfer quality are best assessed through human evaluation in addition to automated metrics. To this end, we conducted a user study comparing several state-of-the-art models, including our own (\textit{DisMo}), to evaluate their performance in terms of realism, prompt adherence, and motion transfer quality. As shown in \cref{tab:motion_transfer_results}, while MotionDirector achieves high motion transfer quality (25.96\%), it exhibits a notable decline in prompt adherence (7.47\%), likely due to visual inconsistencies and content bleeding. In contrast, our method, \textit{DisMo}, attains comparable motion transfer quality (24.71\%) while substantially outperforming MotionDirector in both realism and prompt adherence. Moreover, \textit{DisMo} achieves these results with markedly higher inference efficiency than existing methods. We refer the reader to Section~E in the supplementary for further details.

\begin{table}[h!]
\centering
\caption{\textbf{Human Evaluation of Motion Transfer.} We compare with four other motion transfer methods with respect to realism, prompt matching, and motion transfer quality. \textit{DisMo} achieves the highest scores in realism and prompt matching by a wide margin, while remaining competitive in motion transfer quality.}
\adjustbox{max width=\columnwidth}{
\begin{tabular}{lccc}
\toprule
\textbf{Method} & \textbf{Realism (\%)} & \textbf{Prompt Matching (\%)} & \textbf{Motion Transfer Quality (\%)} \\
\midrule
DMT~\citep{yatim2024space} & 10.93 & 9.60 & 17.73 \\
MotionDirector~\citep{zhao2024motiondirector} & 10.98 & 7.47 & \textbf{25.96} \\
VMC~\citep{jeong2024vmc} & \underline{20.04} & \underline{26.13} & 16.98 \\
MotionClone~\citep{ling2024motionclone} & 19.91 & 19.42 & 14.62 \\
\textit{DisMo (Ours)} & \textbf{38.13} & \textbf{37.38} & \underline{24.71} \\
\bottomrule
\end{tabular}
}
\label{tab:motion_transfer_results}
\end{table}

\subsection{Appearance and Structural Disentanglement}\label{subsec:disentanglement}
\paragraph{Identity Reconstruction from Motion Representations.} To assess the extent to which our motion representations disentangle person-specific appearance from action dynamics, we designed a controlled identity classification experiment. Ideally, motion representations should capture dynamic patterns while minimizing appearance information. To evaluate this property, we utilize the Invariant Action Recognition Dataset (IARD) \citep{DVN/DMT0PG_2019}, which contains controlled recordings of five actors performing five different actions from multiple viewpoints. We extract features from videos of different actors performing a subset of actions and evaluate identity classification on a disjoint action excluded from the training set, enabling analysis of identity-invariant motion representations. A $k$-Nearest Neighbors classifier ($k=20$) was used to predict the identity of the actor based solely on the extracted representations. Higher classification accuracy indicates greater retention of appearance-specific information, while lower accuracy suggests stronger invariance to identity. We compare our method against VideoMAE, VideoMAEv2, and V-JEPA in \cref{tab:identity_classification}. The results reveal clear differences in the amount of identity information each model retains: Our method achieves 23.82\% accuracy, close to the 20\% random baseline for five-way classification, indicating minimal encoding of identity features. In contrast, VideoMAEv2 reaches 58.94\%, while V-JEPA and VideoMAE achieve 96.23\% and 99.14\%, respectively. These findings suggest that our approach disentangles motion from appearance more effectively, making it particularly well-suited for identity-invariant motion understanding and transfer.

We further ablate action classification accuracy on IARD in a similar setting. Specifically, we set up a $k$NN-based action classification experiment and adopt an identity-based split: videos from one individual are held out for testing, while videos from the remaining four individuals are used for training. This evaluation protocol offers a more realistic and challenging assessment by requiring generalization across identities, thereby measuring a model’s ability to capture action dynamics independently of person-specific cues. As shown in \cref{tab:identity_classification}, our representations achieve the best performance on this task, demonstrating that they are both structured and semantically meaningful.

\begin{table}[htbp]
    \centering
    \caption{\textbf{Identity Disentanglement.} We evaluate disentanglement from appearance by performing identity classification on IARD, which contains annotations for both actions and identities. Low identity classification performance indicates high disentanglement. Our motion representations encode significantly less appearance information, while performing best on action classification.}
    \vspace{3mm}
    \label{tab:identity_classification}
    \small
    \begin{tabular}{lccc}
        \toprule
        \textbf{Model} & \textbf{Action Accuracy  $\uparrow$} & \textbf{Identity Accuracy $\downarrow$} & \textbf{vs. Random (20\%)} \\
        \midrule
        VideoMAE~\citep{tong2022videomae} & 73.44 & 99.14 & +79.14 \\
        VideoMAEv2~\citep{wang2023videomae} & 80.25 & \underline{58.94} & \underline{+38.94} \\
        V-JEPA~\citep{bardes2024revisitingfeaturepredictionlearning} & \underline{82.03} & 96.23 & +76.23 \\
        \textit{DisMo (Ours)} & \textbf{90.74} & \textbf{23.82} & \textbf{+3.82} \\
        \bottomrule
    \end{tabular}
\end{table}

\paragraph{Appearance and Structure Invariance during Motion Transfer.} We further evaluate the invariance of motion representations under appearance and structural changes in a motion transfer setting. Starting with an original driving video, we construct two augmented versions: one with photometric transformations (e.g., changes in brightness, contrast, and color) and one with geometric transformations (e.g., rotations, cropping, and spatial shifts). We then perform motion transfer using all three driving videos, the original and the two augmented versions, while keeping the same source prompt and seed. Ideally, an invariant motion representation should produce minimal differences between these outputs. Hence, we compare the generated videos driven by the two augmented views against the one generated using the unaltered video. We compute the CLIP distance~\citep{radford2021learning} as a semantic, SSIM~\citep{wangzhou2004image} as a structural, LPIPS~\citep{zhang2018unreasonable} as a patch-wise as well as the L1 distance as a pixel-wise similarity measurement. As shown in \cref{tab:identity_classification}, our method demonstrates significantly higher invariance towards these transformations compared to other motion transfer methods on all metrics, indicating a high degree of appearance and structural disentanglement. We also notice the positive effect that the augmentations provide during training. While a model trained without augmentations nevertheless demonstrates substantially higher invariance compared to other baselines, a model trained with augmentations performs notably better, especially with regard to geometric transformations.

\begin{table}[h!]
\centering
\caption{\textbf{Generation Similarity under Driving Video Augmentations.} We measure invariance between generated videos driven by unaltered and augmented input videos using both photometric and geometric transformations. \textit{DisMo} showcases a significantly higher degree of invariance towards these alternations compared to the baselines. This effect is reinforced when applying augmentations during training.}
\adjustbox{max width=\columnwidth}{
\begin{tabular}{lcccccccc}
\toprule
\textbf{Method} & \multicolumn{4}{c}{\textbf{Photometric Augmentations}} & \multicolumn{4}{c}{\textbf{Geometric Augmentations}} \\
\cmidrule(lr){2-5} \cmidrule(lr){6-9}
 & \textbf{CLIP} $\uparrow$ & \textbf{SSIM} $\uparrow$ & \textbf{L1} $\downarrow$ & \textbf{LPIPS} $\downarrow$ & 
   \textbf{CLIP} $\uparrow$ & \textbf{SSIM} $\uparrow$ & \textbf{L1} $\downarrow$ & \textbf{LPIPS} $\downarrow$ \\
\midrule
DMT~\citep{yatim2024space} & 0.66 & 0.51 & 0.17 & 0.59 & 0.66 & 0.48 & 0.18 & 0.62 \\
MotionDirector~\citep{zhao2024motiondirector} & 0.62 & 0.46 & 0.15 & 0.55 & 0.60 & 0.44 & 0.16 & 0.60 \\
VMC~\citep{jeong2024vmc} & 0.71 & 0.44 & 0.17 & 0.58 & 0.71 & 0.40 & 0.20 & 0.62 \\
MotionClone~\citep{ling2024motionclone} & 0.76 & 0.52 & 0.13 & 0.46 & 0.72 & 0.43 & 0.18 & 0.58 \\
\textit{DisMo (w/o augmentations)} & \underline{0.89} & \underline{0.70} & \underline{0.09} & \underline{0.26} & \underline{0.88} & \underline{0.65} & \underline{0.11} & \underline{0.33} \\
\textit{DisMo (Ours)} & \textbf{0.90} & \textbf{0.72} & \textbf{0.08} & \textbf{0.25} & \textbf{0.90} & \textbf{0.71} & \textbf{0.08} & \textbf{0.26} \\
\bottomrule
\end{tabular}
}
\label{tab:augmentation_invariance}
\end{table}

\subsection{Action Classification}

\textit{DisMo}'s motion embeddings, while particularly useful for generative motion transfer tasks, demonstrate significant generalization potential for discriminative downstream applications due to their abstract representation of dynamic properties. To investigate their generalization and effectiveness, we evaluate \textit{DisMo} embeddings on a discriminative task, namely action classification.

The strengths of \textit{DisMo}'s motion embeddings are particularly apparent in zero-shot action classification tasks on datasets where motion is a primary discriminative cue. We benchmark our method against state-of-the-art video representation learning approaches on datasets that emphasize temporal dynamics and motion patterns, making them ideal for evaluating motion-centric representations. Specifically, we evaluate on ARID~\citep{xu2021arid}, which tests robustness under low-light conditions; IARD~\citep{DVN/DMT0PG_2019}, designed to assess identity-invariant motion understanding; Jester~\citep{materzynska2019jester}, which comprises short hand gesture clips for evaluating fine-grained local motion sensitivity; and Something-Something V2~\citep{goyal2017something}, a large-scale dataset focused on nuanced temporal interactions.

\begin{table}[htbp]
    \centering
    \caption{\textbf{Zero-shot Action Classification.} We compare \textit{DisMo} with state-of-the-art baselines in frozen evaluation
    with a $k$NN probe on multiple downstream action classification datasets. Extracted features are mean-pooled before being passed to the $k$NN classifier. All models
    are evaluated at resolution $224^2$ using a single clip from the video. Compared to other video baselines, DisMo exhibits a consistent improvement across all downstream datasets, showcasing the superiority of abstract motion representations on zero-shot frozen evaluation over competing video representations.}
    \vspace{3mm}
    \label{tab:retrieval_accuracy_results}
    \small
    \adjustbox{max width=\linewidth}{
    \begin{tabular}{lccccccc}
        \toprule
        \textbf{Method} & \textbf{Arch.} & \textbf{Params.} & \textbf{Samples seen} & \textbf{ARID} $\uparrow$ & \textbf{Jester} $\uparrow$ & \textbf{SSv2} $\uparrow$ & \textbf{IARD} $\uparrow$ \\
        \midrule
        VideoMAE~\citep{tong2022videomae} & ViT-L/16 & 343M & 410M & 17.29 & 20.11 & 7.06 & 73.44 \\
        VideoMAE v2~\citep{wang2023videomae} & ViT-L/16 & 304M & 1.08B & \underline{32.61} & \underline{43.83}& 16.56 & 80.25 \\
        V-JEPA~\citep{bardes2024revisitingfeaturepredictionlearning} & ViT-L/16 & 200M & 270M & 25.16 & 30.84 & \underline{21.11} & \underline{82.03} \\
        \textit{DisMo (Ours)} & DisMo-B & 172M & 17M${}^\dagger$ & \textbf{57.29} & \textbf{56.66} & \textbf{22.19} & \textbf{90.74} \\
        \bottomrule
    \end{tabular}
    }
    {\scriptsize ${}^\dagger$ The number of samples reported here refers only to video clips and does not include images used to pre-train the DINOv2 and DiT backbones.}
\end{table}

Our model demonstrates strong performance across a variety of zero-shot action classification benchmarks, with a notable lead in the k-NN setting (see \cref{tab:retrieval_accuracy_results}). On datasets like ARID, Jester, and SSv2, which all require understanding temporal motion patterns and generalize across diverse visual conditions, DisMo consistently outperforms state-of-the-art baselines. For instance, \textit{DisMo} improves k-NN accuracy by a large margin on ARID (+24.7 points over V-JEPA) and achieves the highest scores on Jester and SSv2, indicating robust motion-sensitive representations.
In the identity-conditioned IARD setting, \textit{DisMo} achieves the highest k-NN accuracy (90.74\%), outperforming all baselines, reflecting a cleaner, less entangled representation, without any supervision or adaptation.

Our identity reconstruction analysis (Section~\ref{subsec:disentanglement}) further supports this interpretation. We observed that \textit{DisMo} encodes substantially less identity information than other models, which often reach near-perfect identity reconstruction. This property of disentangling motion from appearance enables better generalization across individuals and tasks, as the representations focus on action dynamics rather than performer-specific features. These results collectively underscore the promise of motion-centric representation learning for building generalizable and robust video models.

\section{Conclusion}
In this work, we introduce \textit{DisMo}, a novel paradigm for learning abstract motion representations from video data using a purely reconstruction-based objective. We demonstrate that these representations are inherently decoupled from appearance, making them well-suited for motion transfer. \textit{DisMo} outperforms existing open-world transfer methods on qualitative and quantitative evaluations. Furthermore, we demonstrate that the learned embeddings are also effective for downstream tasks such as action classification, outperforming state-of-the-art video representation models in a zero-shot setting. While our analysis focuses on zero-shot probing, exploring more advanced techniques such as linear or attentive probing could offer deeper insights into the structure and capabilities of the learned motion space, which we leave for future work.

\clearpage

\section*{Acknowledgements}
This project has been supported by the project “GeniusRobot” (01IS24083) funded by the Federal Ministry of Research, Technology and Space (BMFTR), the bidt project KLIMA-MEMES, the German Federal Ministry for Economic Affairs and Energy (BMWE) within the project “NXT GEN AI METHODS – Generative Methoden für Perzeption, Prädiktion und Planung”, and the Horizon Europe project ELLIOT (GA No. 101214398). The authors gratefully acknowledge the Gauss Center for Supercomputing for providing compute through the NIC on JUWELS/JUPITER at JSC and the HPC resources supplied by the NHR@FAU Erlangen.

\bibliography{main}
\appendix
\setcounter{figure}{0}
\setcounter{table}{0}
\setcounter{equation}{0}
\setcounter{section}{0}
\renewcommand\thefigure{\Alph{figure}}
\renewcommand\thetable{\Alph{table}}
\renewcommand\thesection{\Alph{section}}

\newpage

\section{Broader Societal Impact} While this work aims to advance realistic motion transfer with potential applications in digital storytelling, entertainment, and educational purposes, such progress also carries risks of misuse to generate harmful content without consent. We emphasize transparency in datasets and evaluation, and encourage ongoing efforts to address biases and promote responsible deployment.

\section{Limitations}
Our work is limited by the capabilities of the flow-matching-based decoder, which, while efficient, may struggle to generate high-fidelity or semantically rich images in complex scenarios. Additionally, the performance is constrained by the distribution and inherent biases of the training dataset, which can affect generalization, fairness, and robustness to out-of-distribution samples.

\section{Implementation Details}

\subsection{Pretraining details}
\textbf{Architectures} We use DINOv2-B as our frame embedder. As our sequence embedder we use a 3D ViT-B/16 with RoPe positional embeddings.
As our frame generator, we use a DiT-XL.
The frame embedder takes as input a short video clip of 8 frames, sampled at 6 fps.

\begin{table}[ht]
\centering
\caption{\textbf{Pretraining hyper-parameters for our model.}}
\label{tab:pretraining_base}
\small
\setlength{\tabcolsep}{6pt}
\begin{tabular}{llc}
\toprule
\textbf{Category} & \textbf{Hyper-parameter} & \textbf{Value} \\
\midrule
\multirow{4}{*}{\textit{data}} 
  & datasets & K-710/SSv2/MiT/OpenVid-1m \\
  & resolution & $256 \times 256$ \\
  & num\_frames & 8 \\
   & fps & 6 \\
\midrule
\multirow{6}{*}{\textit{geometric augmentations}} 
  & scale & (0.75, 1.0) \\
  & translate & (-0.3, 0.3) \\
  & angle & (-30, 30) \\
  & shear & (-15, 15) \\
  & horizontal flip & \checkmark \\
  & padding mode & border \\
\midrule
\multirow{4}{*}{\textit{photometric augmentations}} 
  & brightness & (0.5, 1.5) \\
  & contrast & (0.5, 1.5) \\
  & hue & (-0.3, 0.3) \\
  & saturation & (0.5, 1.5) \\
\midrule
\multirow{7}{*}{\textit{optimization}} 
  & batch\_size & 32 \\
  & total number of iterations & 530k \\
  & warmup iterations & 5000 \\
  & lr scheduler & constant w/ warmup \\
  & lr & $10^{-4}$ \\
  & AdamW $\beta$ & $(0.9, 0.95)$ \\
\midrule
\multirow{6}{*}{\textit{architecture}} 
  & frame\_embed\_depth & 12 \\
  & frame\_embed\_dim & 768 \\
  & sequence\_embed\_depth & 12 \\
  & sequence\_embed\_dim & 768 \\
  & frame\_generator\_depth & 28 \\
  & frame\_generator\_dim & 1152 \\
\midrule
\multirow{2}{*}{\textit{hardware}} 
  & dtype & bfloat16 \\
  & accelerator & GH200 96G \\
\bottomrule
\end{tabular}
\end{table}

\textbf{Training Data}
To cover a wide variety of real-world videos, we train on a mixture of datasets covering over videos. This mixture is composed of Kinetics-710~\citep{kay2017kinetics}, which combines Kinetics-400, 600 and 700, Something-Something-v2 (SSv2)~\citep{goyal2017something}, Moments in Time (MiT)~\citep{monfort2019moments}, and OpenVid1M~\citep{nan2025openvidm}. We show an overview of the size and domain of the various datasets in \cref{tab:training_dataset_composition}. Overall, this dataset composition is similar to that used by V-JEPA~\citep{bardes2024revisitingfeaturepredictionlearning}, which among others combines Kinetics-710 with Something-Something-v2. We further extend this mixture with the high-quality OpenVid-1M \citep{nan2025openvidm} dataset, which is more suitable for generative tasks compared to the other datasets.

\begin{table}[htb]
    \centering
    \adjustbox{max width=\linewidth}{
        \begin{tabular}{lccc}
            \toprule
            Dataset & Domain & \# Clips & Duration [h] \\
            \midrule
            Kinetics-710~\citep{kay2017kinetics} & human-centered & 0.65M & 1.8k \\
            Something-Something-v2~\citep{goyal2017something} & open & 108k & 121 \\
            Moments in Time~\citep{monfort2019moments} & open & 1M & 833 \\
            OpenVid-1M~\citep{nan2025openvidm} & open & 1M & 2.1k \\
            \midrule
            \textit{Combined} & open & 2.8M & 4.9k \\
            \bottomrule
        \end{tabular}
    }
    \caption{Training Data Composition.}
    \label{tab:training_dataset_composition}
\end{table}

\paragraph{Kinetics-710 Overlaps} There are significant overlaps between train and evaluation splits of the various subsets (K-400, K-600, K-700) that comprise the Kinetics-710 dataset. To enable fair comparisons, we remove any clips present in any of the three validation or test sets from our training dataset. We also de-duplicate the remaining training dataset by their respective YouTube video ids.

\subsection{Evaluation details}

\textbf{Datasets} Something-Something v2 (SSv2) \citep{goyal2017something} is a large dataset of human-object interactions designed to test fine-grained motion understanding while minimizing reliance on background or object identity. Kinetics-400 \citep{kay2017kinetics} contains 240,000 YouTube clips labeled with 400 action classes and serves as a general-purpose benchmark for appearance-driven video understanding. ARID \citep{xu2021arid} (Action Recognition in the Dark) includes 5,500 RGB-D videos captured in low-light settings to evaluate robustness to visibility degradation. Jester \citep{materzynska2019jester} contains over 148,000 short clips of hand gestures, providing a benchmark for evaluating fine-grained, local temporal motion sensitivity. IARD \citep{DVN/DMT0PG_2019} (Invariant Action Recognition Dataset) includes controlled recordings of five actors performing five actions from multiple viewpoints, enabling analysis of identity- and viewpoint-invariant motion recognition. HMDB51 \citep{Kuehne11} is a widely used benchmark for human action recognition, consisting of $\sim$7,000 video clips distributed across 51 action categories, collected from various sources including movies and YouTube, with diverse scene complexity and motion dynamics.

\textbf{LTX Finetuning details} We adapt both the attention and feedforward layers of LTX using conditional LoRA modules with a rank of 64. Each LoRA module receives a motion sequence as input. We use a batch size of 64 for training. For optimization, we use the AdamW optimizer with $\beta_1 = 0.9$ and $\beta_2 = 0.99$, a learning rate of $5 \times 10^{-5}$, weight decay of $0.0001$, and training is performed with 32-bit precision.

LTX operates on high frame rate videos and temporally downsamples them by a factor of 8 using its VAE. In contrast, our motion extractor processes videos at a lower frame rate, taking only 8 frames as input. To ensure alignment between the motion sequence and the LTX latent representations, we apply the following strategy:

\begin{itemize}
    \item We sample a video at approximately 24 fps with a total of 29 frames, where the first frame serves as the source frame.
    \item These 29 frames are chosen such that we can extract 8 frames spaced 4 frames apart: the source frame (frame 0), followed by frames 4, 8, 12, 16, 20, 24, and 28. This satisfies the equation:
    \[
    1 + 7 \times 4 = 29
    \]
    \item LTX downsamples this video by encoding the source frame and every 8\textsuperscript{th} frame thereafter (frames 0, 8, 16, 24) into a latent representation.
    \item To align motion embeddings with these latents, we extract every 4\textsuperscript{th} frame (frames 0, 4, 8, 12, 16, 20, 24, 28) and pass them through our motion extractor.
    \item For each pair of consecutive motion embeddings (except for the first one, which remains a singular embedding), we concatenate them to form a combined embedding. This results in a motion sequence of 4 embeddings aligned with the 4 LTX latents.
    \item This sequence is then transformed by a mapping network and used to condition the LoRA modules. The mapping network consists of two feedforward layers, each applying RMS normalization at the beginning and end, and using a Linear-GEGLU \citep{shazeer2020gluvariantsimprovetransformer} activation. Each feedforward layer expands the dimensionality by a factor of 3 and subsequently reduces it back, operating at a width of 768.

\end{itemize}

\textbf{Identity Reconstruction from Motion Representations} We provide additional details on the identity classification experiment on the IARD dataset. We selected videos of 5 individuals and used only four actions (\texttt{['eat', 'run', 'walk', 'jump']}) for training, holding out the \texttt{'drink'} action for testing. Motion representations were extracted using VideoMAE, VideoMAEv2, and Video JEPA. Videos were processed at 6 FPS with 16-frame windows, resized to 224$\times$224, and uniformly preprocessed across models. Temporal features were averaged to produce a single embedding per video. A k-NN classifier ($k=20$) was trained to predict identity based on the training actions and evaluated on the held-out \texttt{'drink'} action to assess the extent of identity information encoded in the representations.

\textbf{IARD identity-based split for action classification} In this experiment on the Invariant Action Recognition Dataset (IARD), we used an identity-based split to better test the generalization of motion representation models across individuals. The dataset includes 5 people performing 5 actions (\texttt{walk}, \texttt{run}, \texttt{jump}, \texttt{drink}, \texttt{eat}). We trained models on videos from four individuals (Andrea, Leyla, Gu, Georgios) and tested on videos from the fifth (Steve), ensuring no overlap in identity between train and test. This split contrasts with standard random splits and provides a more realistic measure of how well models capture action dynamics independently of personal appearance cues.

\subsection{Ablation Study}
During training, we introduce several key strategies to enhance disentanglement between motion and appearance in our model. Additionally, we apply dropout to motion and frame conditioning, which serves a dual purpose: it stabilizes the training process and allows for subsequent evaluation of the frame-unconditional reconstructor. This evaluation is conducted by reconstructing frames solely from motion embeddings, without previous frame conditioning.

To comprehensively measure performance during ablation, we employ three complementary metrics: LPIPS \citep{zhang2018unreasonable}, CLIP \citep{radford2021learning} score, and Mutual Information Ratio (MIR).  Ideally, if no appearance information leaks into the motion embeddings, we expect poorer reconstructions, indicated by higher LPIPS scores and lower CLIP similarities. We use a subset of the HMDB-51 dataset for LPIPS and CLIP, and a subset of the IARD dataset for MIR.

With LPIPS we quantify the perceptual similarity with respect to visual realism, whereas with CLIP we measure the overall semantic similarity between the reconstructed frame and the original one. MIR provides a formal quantification of disentanglement by comparing the mutual information between the motion embeddings and two distinct factors -- actions and identities. Formally, we define MIR as:
\begin{equation}
\text{MIR} = \frac{I(z; c_{\text{action}})}{I(z; c_{\text{identity}})},
\end{equation}

where $c_{\text{action}}$ and $c_{\text{identity}}$ are the respective labels for embeddings $z$ and the mutual information $I(\cdot; \cdot)$ measures the shared information between variables, defined as:
\begin{equation}
I(z; y) = H(y) - H(y \mid z),
\end{equation}

with entropy $H(\cdot)$ quantifying uncertainty.
We estimate entropy using $k$-NN ($k=5$) distances, as described in \citep{kraskov2004estimating} and \citep{10.1371/journal.pone.0087357}.

Table \ref{tab:ablations} summarizes the outcomes of our ablation study. Starting from a baseline model without conditioning or augmentations, adding previous frame conditioning significantly prevents appearance leaking and improves disentanglement. Integrating augmentations further boosts these improvements, especially regarding disentanglement.
Collectively, these results underscore the efficacy and necessity of each component in achieving robust disentanglement between motion and appearance.

\begin{table}[ht]
    \centering
    \begin{minipage}[t]{0.48\textwidth}
        \centering
        \caption{\textbf{Ablation Study.} Comparison of model variants with and without previous frame conditioning and augmentations.
        }
        \label{tab:ablations}
        \small
        \adjustbox{max width=\linewidth}{
        \begin{tabular}{lccc}
            \toprule
            \textbf{Model} & \textbf{LPIPS $\uparrow$} & \textbf{CLIP $\downarrow$} & \textbf{MIR $\uparrow$} \\
            \midrule
            Baseline & 0.47 & 0.81 & 0.47 \\
            + Previous frame & 0.72 & 0.65 & 3.07 \\
            + Augmentations (ours) & 0.76 & 0.65 & 5.56 \\
            \bottomrule
        \end{tabular}
        }
    \end{minipage}%
    \hfill
    \begin{minipage}[t]{0.48\textwidth}
        \centering
        \caption{\textbf{Disentanglement.} Comparison of appearance disentanglement performance between our model and V-JEPA.
        }
        \label{tab:disentanglement}
        \small
        \begin{tabular}{lccc}
            \toprule
            \textbf{Model} & \textbf{LPIPS $\uparrow$} & \textbf{CLIP $\downarrow$} & \textbf{MIR $\uparrow$} \\
            \midrule
            V-JEPA & 0.70 & 0.70 & 3.72 \\
            Ours & \textbf{0.76} & \textbf{0.65} & \textbf{5.56} \\
            \bottomrule
        \end{tabular}
    \end{minipage}
\end{table}

To enable a direct performance comparison with V-JEPA, we trained an identical reconstructor using the same setup as our model, with the motion extractor replaced by a frozen V-JEPA. As shown in \cref{tab:disentanglement}, our model achieves superior disentanglement.

We additionally ablate \textit{DisMo} with other high-quality video generation backbones, namely SparseCtrl~\citep{guo2023sparsectrladdingsparsecontrols} and CogVideoX-5B~\citep{yang2025cogvideoxtexttovideodiffusionmodels}. As shown in \cref{tab:video_quality}, using a more capable generator like CogVideoX-5B leads to improved motion fidelity and temporal consistency, without any modification to the motion embeddings produced by \textit{DisMo}. This demonstrates that \textit{DisMo} is generally applicable to off-the-shelf video models without being constrained to a specific one. We further assessed the quality and realism of the generated videos using FID and FVD scores, observing consistent improvements with stronger video backbones. This demonstrates that \textit{DisMo} benefits from more powerful architectures and remains compatible with future models. For this evaluation, we generate 2,000 videos per model, each 25 frames long, yielding 50,000 frames in total for FID computation. Motion is transferred from randomly sampled videos and prompts from the Koala-36M~\citep{wang2025koala36mlargescalevideodataset} dataset.

\begin{table}[h!]
\centering
\caption{\textbf{Ablating Video Generation Backbones for Motion Transfer.} Effect of different video generation backbones on temporal consistency, motion fidelity, and visual quality metrics (FID, FVD) of generated motion transfers. Using a stronger generator like CogVideoX improves motion fidelity and temporal consistency without altering DisMo’s motion embeddings.} %
\adjustbox{max width=\columnwidth}{
\begin{tabular}{lcccc}
\toprule
\textbf{Model} & \textbf{Temporal Consistency}~$\uparrow$ & \textbf{Motion Fidelity}~$\uparrow$ & \textbf{FID}~$\downarrow$ & \textbf{FVD}~$\downarrow$ \\
\midrule
LTX~\citep{HaCohen2024LTXVideo} & 0.95 & 0.75 & 88.54 & 298.4 \\
SparseCtrl~\citep{guo2023sparsectrladdingsparsecontrols} & 0.95 & 0.73 & 116.8 & 327.9 \\
CogVideoX-5B~\citep{yang2025cogvideoxtexttovideodiffusionmodels} & \textbf{0.97} & \textbf{0.78} & \textbf{62.99} & \textbf{206.3} \\
\bottomrule
\end{tabular}
}
\label{tab:video_quality}
\end{table}

\section{Latent Space Analysis}

\subsection{Cluster Visualization}
\begin{figure}[ht]
  \centering
    \includegraphics[width=\linewidth]{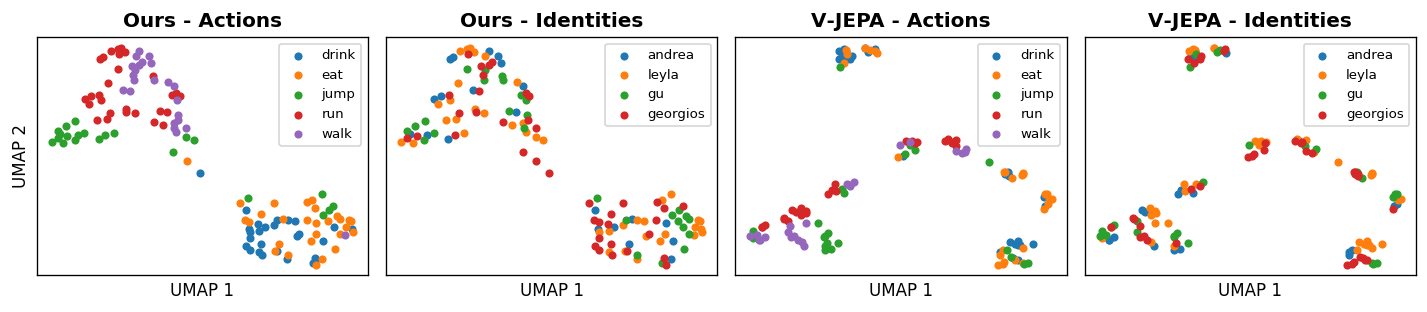}
  \caption{\textbf{UMAP visualization of the IARD dataset.} Compared to V-JEPA, our model shows better grouping by action and almost no grouping by identity.}
  \label{fig:umap}
\end{figure}

We visualize embeddings from both DisMo and V-JEPA on the IARD dataset, which provides action and identity labels for each video sample. Using UMAP for dimensionality reduction and coloring the resulting sample points by either action or identity reveals notable differences between the two models, see \cref{fig:umap}. When colored by action, embeddings from our model form well-defined clusters, while coloring by identity results in no clear grouping—indicating strong sensitivity to motion and minimal association with appearance. In contrast, V-JEPA’s embeddings show little to no clustering by either action or identity. Instead, its clusters primarily reflect viewpoint differences, consistent with the five distinct camera angles present in the dataset.

\subsection{Cycle Identification}
\begin{figure}[ht]
  \centering
    \includegraphics[width=0.8\linewidth]{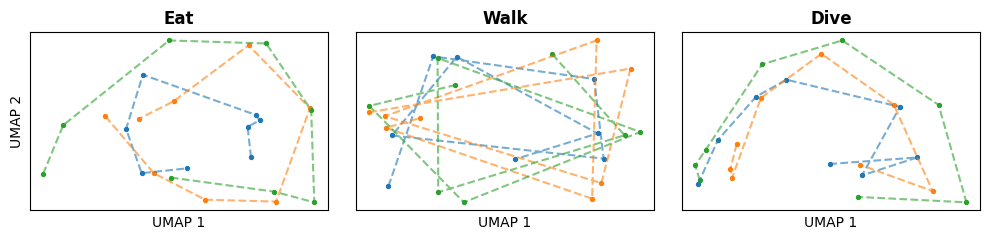}
  \caption{\textbf{PCA cycle visualization of three different classes from the IARD dataset.} Three videos per class, indicated by different colors, showing clear cycles in our model's latent representations.}
  \label{fig:cycle}
\end{figure}

\cref{fig:cycle} shows PCA visualizations of individual videos from three different classes in the IARD dataset: eat, walk, and run. The repetitive nature of these actions produces distinct cyclic patterns in the latent space. Slower movements, such as eating, yield smoother and more continuous trajectories, while faster actions like walking result in less smooth, more abrupt trajectories. None-repeating motions, such as someone diving into a pool, do not result in cycles. This highlights how differences in motion intensity and cycling movements are captured in our model's latent representations.

\subsection{Reversibility}
\begin{figure}[ht]
  \centering
    \includegraphics[width=\linewidth]{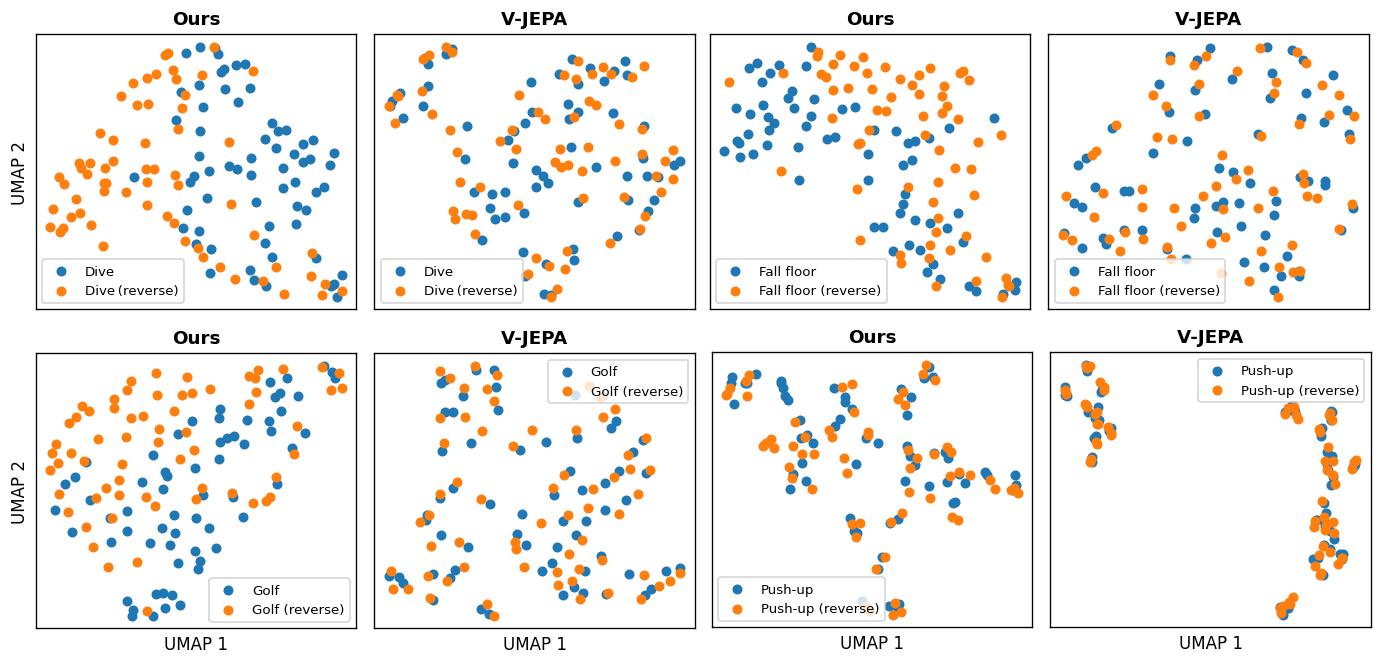}
  \caption{\textbf{UMAP visualizations of reversible motion and irreversible motion on the HMDB-51 dataset.} Compared to V-JEPA, our model shows a better separation for irreversible motion.}
  \label{fig:reversibility}
\end{figure}

A task that is solely based on motion, where appearance provides no useful cues, is determining whether a video is played forward or backward. In such cases, temporal directionality must be inferred from the dynamics of the scene alone.
To test this, we conducted an experiment where each video sample is paired with a temporally reversed version. We then visualize the latent representations using UMAP. As shown in \cref{fig:reversibility}, our model separates forward and reversed sequences in the latent space, but only when the action is inherently irreversible, such as diving into a pool. In contrast, when the motion is reversible e.g. someone doing push-ups, no meaningful separation emerges.
V-JEPA, on the other hand, fails to distinguish forward and reversed sequences in both reversible and irreversible cases. Its latent space shows no significant separation, suggesting its representations are largely appearance biased.

\section{Composite Motion Analysis}
We designed a targeted experiment to evaluate the composability of DisMo’s motion embedding space. To assess which aspects of motion are encoded in the embeddings, we use the YUP++ dataset \citep{feichtenhofer2017temporal}, which contains 600 static videos without camera motion. We then construct three types of video clips:

\begin{enumerate}
    \item \textbf{Object Motion Only.} A randomly sampled clip from the dataset, containing natural object motion but no camera movement.
    \item \textbf{Camera Motion Only.} A different clip, augmented with temporally smooth geometric transformations simulating camera motion, applied only to the initial frame (thus eliminating object motion).
    \item \textbf{Combined Motion.} The sampled clip with both object motion and the same simulated camera motion applied across all frames.
\end{enumerate}

This setup allows us to isolate the two motion components, camera and object, and test whether embeddings extracted from the combined motion videos contain information from both individual motion types. We sample 3200 of such triplets, each containing the three types of clips mentioned before, and encode each of them using DisMo’s motion encoder. We then estimate the mutual information (MI) between the embeddings of the combined-motion clip (\(\mathbf{M}_{both}\)) and those of the camera-only \(\mathbf{M}_{cam}\)) and object-only \(\mathbf{M}_{obj}\)) clips, respectively. High mutual information would suggest that \(\mathbf{M}_{both}\) shares information with \(\mathbf{M}_{cam}\) and \(\mathbf{M}_{obj}\), which is a key indication of composability.

To quantify this, we use the Kraskov–Stögbauer–Grassberger (KSG) estimator \citep{kraskov2004estimating} with \(k = 5\). To compute statistical significance, we further estimate MI under 100 random permutations (i.e., computing MI between unpaired clips), yielding a null distribution with means of around 0.2 nats and standard deviations of around 0.4. The observed MI values between both (\(\mathbf{M}_{both}\), \(\mathbf{M}_{cam}\)) and (\(\mathbf{M}_{both}\), \(\mathbf{M}_{obj}\)) are depicted in \cref{tab:composite_motion}. They are significantly higher than the null baseline, indicating a strong statistical dependency and thus supporting the hypothesis that DisMo’s latent space is compositional with regard to simultaneous camera and object motion.

\begin{table}[htbp]
    \centering
    \caption{\textbf{Motion composability.} We compute the mutual information (MI) between motion embeddings extracted from videos with both camera and object motion, and embeddings extracted from videos consisting of only camera and object motion, respectively. We observe a high significance compared to a random baseline for both types of motion. Object motion MI is significant even with strong camera motion, albeit decreasing the heavier the added camera motion gets.}
    \vspace{3mm}
    \label{tab:composite_motion}
    \small
    \begin{tabular}{lccc}
        \toprule
        \textbf{MI (nats)} & \textbf{Light Camera Motion} & \textbf{Medium Camera Motion} & \textbf{Heavy Camera Motion} \\
        \midrule
        Camera Motion & 1.37 & 2.25 & 4.40 \\
        Object Motion & 3.59 & 2.97 & 1.57 \\
        \bottomrule
    \end{tabular}
\end{table}

\section{Motion Transfer Efficiency}
We report inference time and number of parameters for DisMo and other baselines during motion transfer in \cref{tab:inference_time}. DMT, VMC, and MotionDirector rely on per-sample fine-tuning or optimization, which significantly decreases inference efficiency. In contrast, our method is an off-the-shelf feed-forward generator. Across all configurations, we report DisMo to be the most efficient approach — except when paired with CogVideoX-5B as the video generation backbone, which increases inference time due to sheer model size. However, it is important to note that using CogVideoX-5B is not required to achieve strong motion transfer results: as evidenced by \cref{tab:general_transfer_quantitative}, DisMo combined with lighter backbones such as LTX-Video-2B already achieves better or comparable motion transfer fidelity.

\begin{table}[h]
\centering
\caption{\textbf{Inference Efficiency Comparison.} Inference time and parameter count of each model on an A100 GPU. Optimization-based baselines (VMC, DMT, MotionDirector) require per-sample fine-tuning, leading to slow inference, while our feed-forward \textit{DisMo} variants enable substantially faster generation across all configurations.}
\begin{tabular}{lcc}
\toprule
\textbf{Model} & \textbf{Inference Time (A100)} & \textbf{\# Params} \\
\midrule
VMC & 10 min / video & 6B \\
DMT & 7.5 min / video & 1.7B \\
MotionClone & 45 sec / video & 0.983B \\
MotionDirector & $\sim$5 min / video & 0.983B \\
DisMo-CogVideo & 2 min / video & 6.072B \\
DisMo-LTX & 30 sec / video & 2.172B \\
DisMo-SparseCtrl & 10 sec / video & 1.115B \\
\bottomrule
\end{tabular}

\vspace{2pt}
\begin{minipage}{0.9\linewidth}
\scriptsize
${}^\dagger$ The reported number of parameters is approximate and refers to the largest component of each model, to the best of our knowledge. Exact values may vary depending on implementation details and auxiliary components.
\end{minipage}
\label{tab:inference_time}
\end{table}

\section{Computational Resources for Experiments}
This section gives an overview of the compute resources required to run the various individual experiments presented in this paper. We primarily use Nvidia GH200 96GB modules in an internal cluster for pre-training and Nvidia H200 141GB in an internal cluster for evaluations. The project used approximately 600 H200-hrs across all experiments, including initial and other explorations that did not directly make it into the paper. In the following, we detail the required resources for each experiment.

\begin{itemize}
    \item Pre-training: 192 H200 hours.
    \item Ablations: $2 \times 96$ H200 hours.
    \item Action classification: 8 H200 hours
    \item Identity classification: 0.66 H200 hours
    \item Motion transfers: 17.2 H200 hours
    \item LTX fine-tuning: 200 H200 hours
\end{itemize}

\section{Licenses of Used Assets}

\subsection{Licences of video datasets}
\begin{itemize}
  \item \textbf{IARD}\\
  License: CC0 1.0.\\
  Terms: Community Norms in \url{https://dataverse.org/best-practices/dataverse-community-norms} as well as good scientific practices expect that proper credit is given via citation.\\
  Source: \url{https://dataverse.harvard.edu/dataset.xhtml?persistentId=doi:10.7910/DVN/DMT0PG}

  \item \textbf{Something-Something v2 (SSv2)}\\
  License: Research Use License (Qualcomm).\\
  Terms: Available for research purposes only. Users must agree to Qualcomm's license terms before accessing the dataset.\\
  Source: \url{https://www.qualcomm.com/developer/software/something-something-v-2-dataset}

  \item \textbf{Jester (Gesture Recognition)}\\
  License: Research Use License (Qualcomm).\\
  Terms: Available for research purposes only. Users must agree to Qualcomm's license terms before accessing the dataset.\\
  Source: \url{https://www.qualcomm.com/developer/software/jester-dataset}

  \item \textbf{ARID (Action Recognition in the Dark)}\\
  License: Creative Commons Attribution 4.0 International (CC BY 4.0).\\
  Terms: Free to use, modify, and redistribute with proper attribution.\\
  Source: \url{https://xuyu0010.github.io/arid.html}

  \item \textbf{OpenVid-1M}\\
  License: Creative Commons Attribution 4.0 International (CC BY 4.0).\\
  Terms: The video samples are collected from publicly available datasets. Users must follow the related licenses Panda, ChronoMagic, Open-Sora-plan, CelebvHQ(Unknow)) to use these video samples.\\
  Source: \url{https://huggingface.co/datasets/nkp37/OpenVid-1M}

  \item \textbf{Kinetics-710}\\
  License: Not explicitly specified.\\
  Terms: Kinetics-710 is a video benchmark built upon Kinetics-400, -600, and -700. Its usage is subject to the licensing terms of these underlying datasets.\\
  Source: \url{https://github.com/open-mmlab/mmaction2/blob/main/tools/data/kinetics710/README.md}

  \item \textbf{Moments in Time}\\
  License: Not explicitly specified.\\
  Terms: Intended for academic research.\\
  Source: \url{http://moments.csail.mit.edu}

  \item \textbf{HMDB51}\\
  License: Creative Commons Attribution 4.0 International (CC BY 4.0).\\
  Terms: Free to use, modify, and redistribute with proper attribution.\\
  Source: \url{https://serre-lab.clps.brown.edu/resource/hmdb-a-large-human-motion-database/}

\end{itemize}

\section{Qualitative Comparisons on Text-based Motion Transfer}
We present additional qualitative results for motion transfer and compare them with the previously introduced baseline methods. As most existing approaches do not support motion transfer from a source image, we restrict our comparison to examples generated solely from text prompts. All experiments were conducted using our adapted LTX model on videos that were not seen during training.

These results further demonstrate the effectiveness of our abstract motion representation in enabling motion transfer across diverse scenarios. Our approach consistently preserves both the high-level motion semantics and the content specified in the text prompt without relying on the structure of the driving video, a trade-off that competing methods often struggle to achieve.

\begin{figure}[ht]
  \centering
\includegraphics[width=\linewidth]{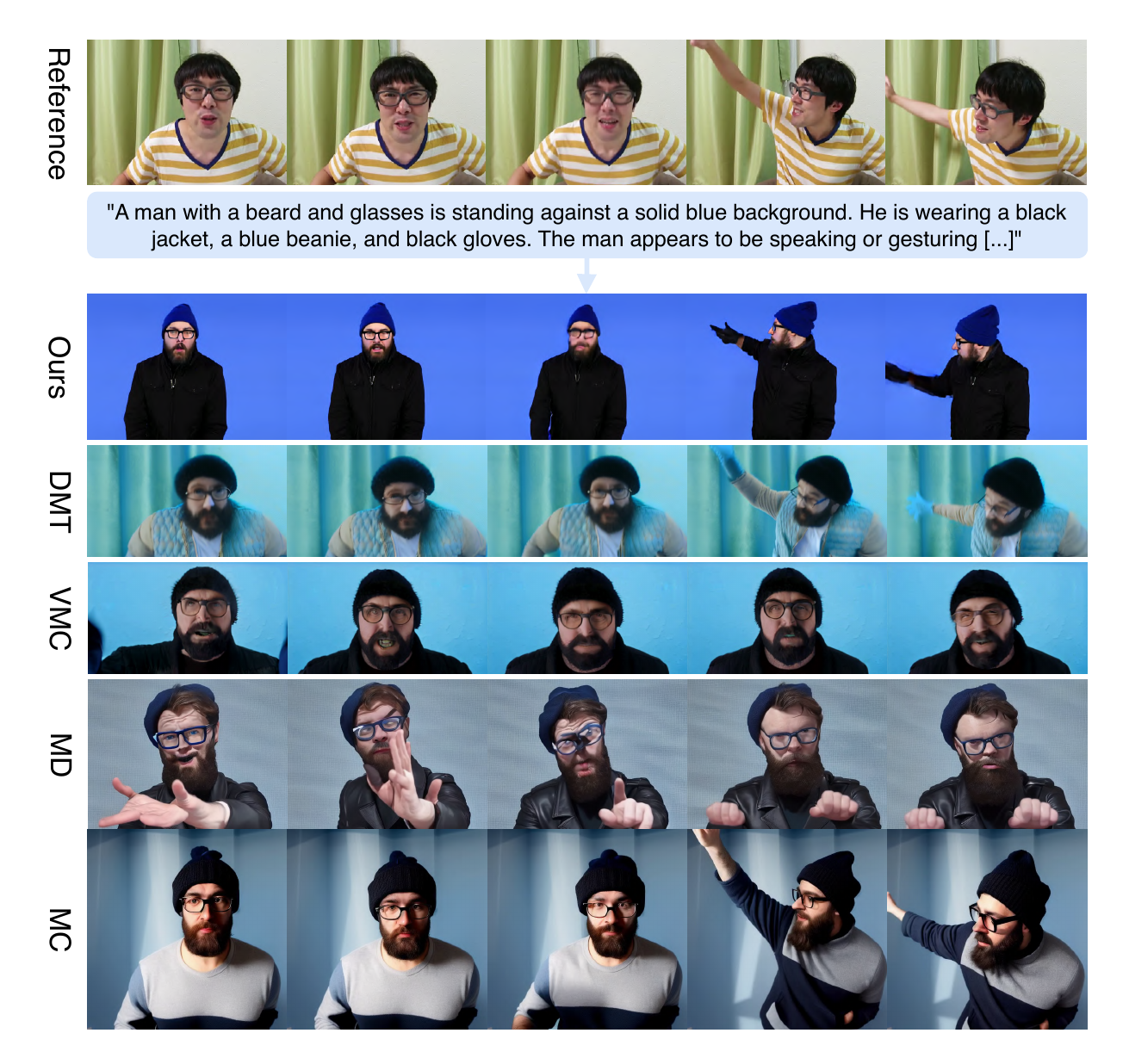}
  \caption{\textbf{Same-category Transfer.} The reference video displays a man leaning towards the camera, rotating his torso and raising his arm. Our method is able to transfer these movements to a new identity with differing appearance, pose, and structure. Other methods either fail to capture the motion completely (\textbf{VMC}, \textbf{MD}) or rely heavily on the source structure (\textbf{DMT}, \textbf{MC}).}
  \label{fig:same_category_1}
\end{figure}

\begin{figure}[ht]
  \centering
\includegraphics[width=\linewidth]{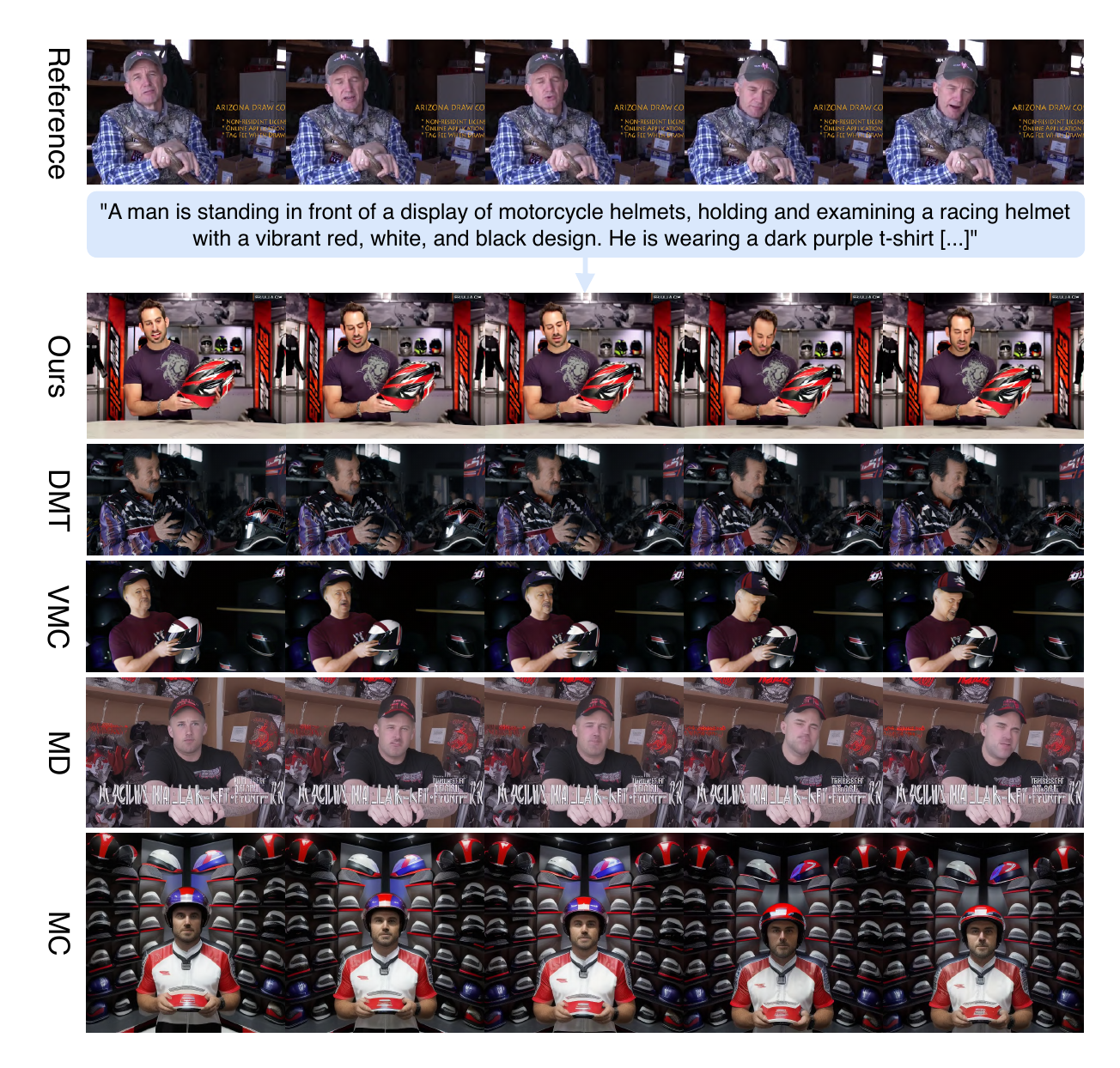}
  \caption{\textbf{Same-category Transfer.} The reference video shows the upper torso of a person,  moving their head and upper body. Our method successfully captures this motion and transfers it to the generated scene, closely following the instructions of the text prompt. Other methods fail to interpret the motion and the input prompt in a meaningful manner, the only exception being \textbf{VMC}.}
  \label{fig:same_category_2}
\end{figure}

\begin{figure}[ht]
  \centering
\includegraphics[width=\linewidth]{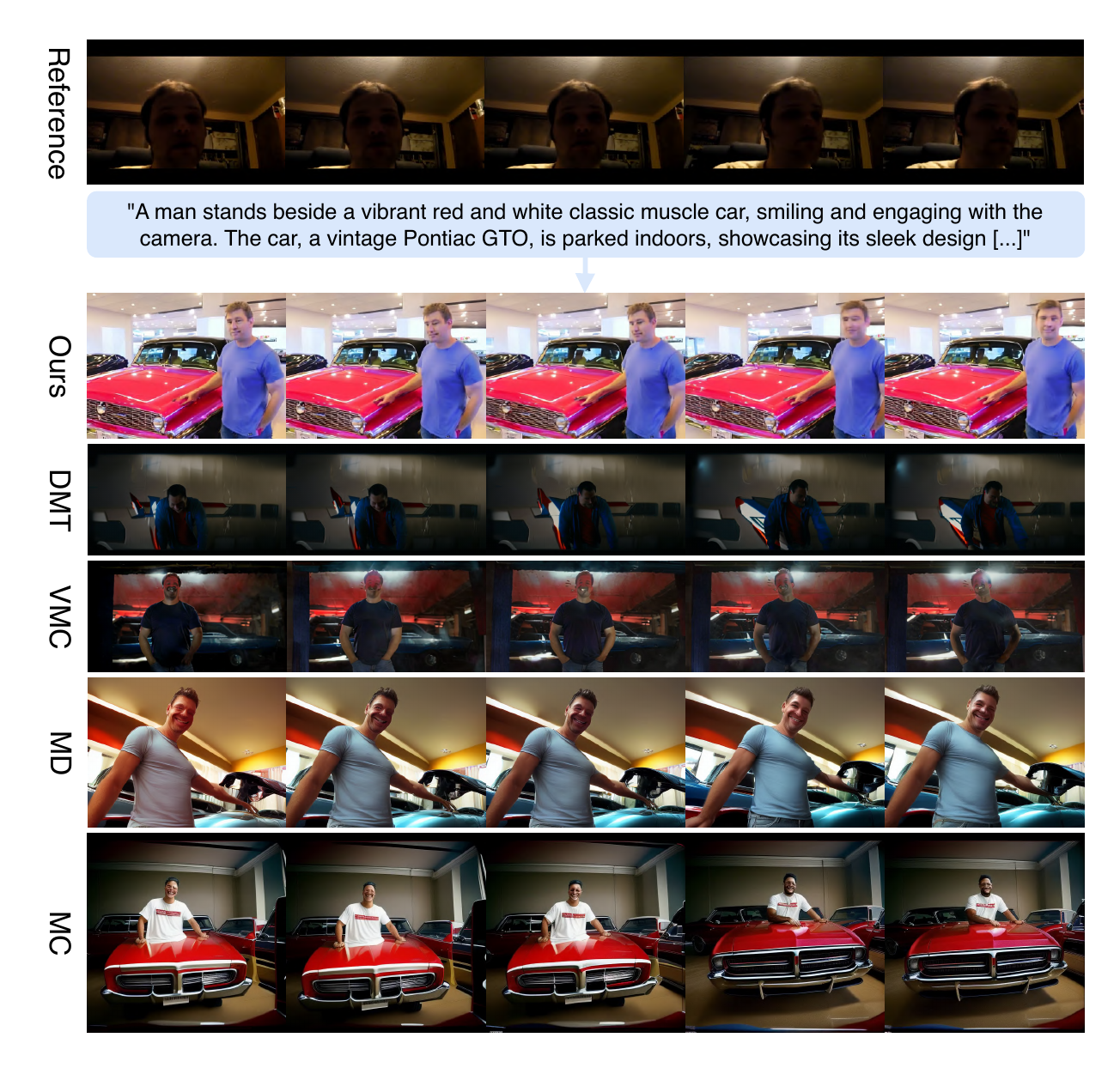}
  \caption{\textbf{Same-category Transfer.} The reference video depicts a talking head in a dark setting, slowly leaning to the left-hand side. Our approach effectively captures the details of the text prompt and accurately attaches the reference head movement to the semantically appropriate part of the target video, despite significant structural differences. In contrast, other methods fail to transfer the reference motion correctly.}
  \label{fig:same_category_3}
\end{figure}

\begin{figure}[ht]
  \centering
\includegraphics[width=\linewidth]{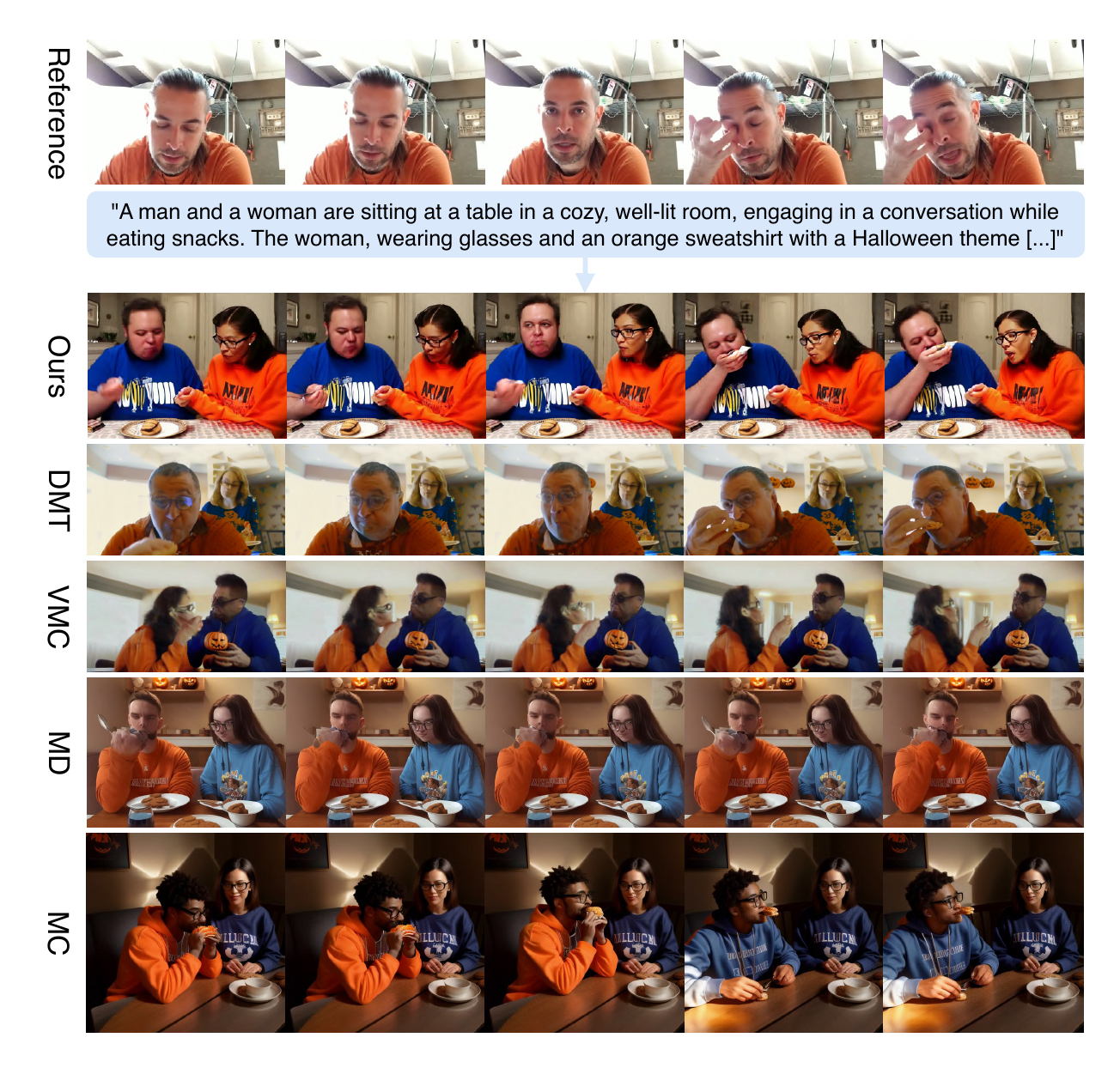}
  \caption{\textbf{Same-category Transfer.} The reference video depicts a close-up shot of a person looking at the camera while performing head and arm movements. Our approach successfully transfers the nuances of this motion to the two target characters and even reinterprets it to represent the action described in the text prompt (eating). In contrast, \textbf{DMT} overly relies on the reference structure, limiting its ability to capture the intended semantics. Furthermore, \textbf{VMC}, \textbf{MD}, and \textbf{MC} fail to capture key nuances of the reference motion, such as the arm movement.}
  \label{fig:same_category_4}
\end{figure}

\begin{figure}[ht]
  \centering
\includegraphics[width=\linewidth]{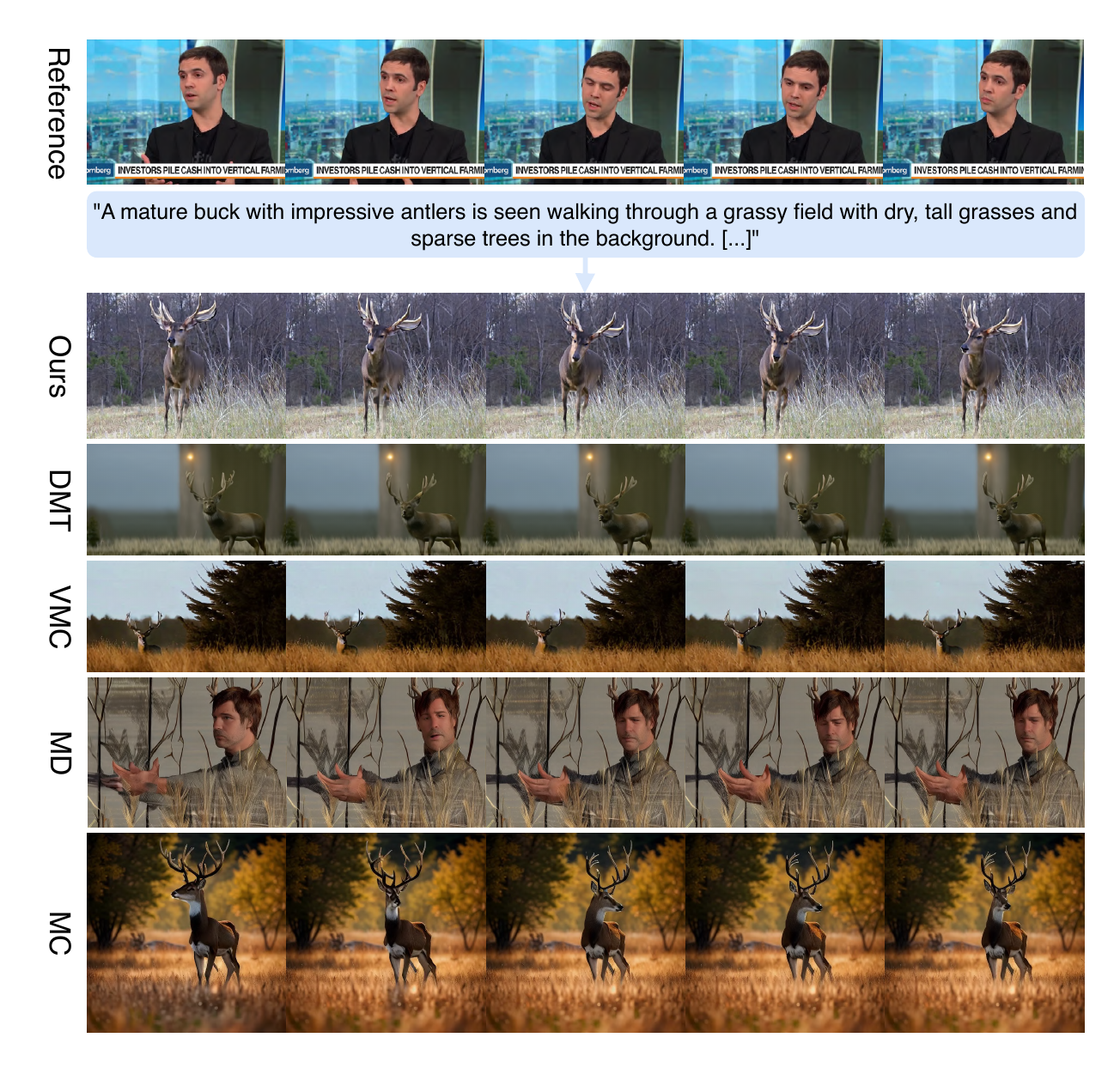}
  \caption{\textbf{Cross-category Transfer.} The reference video shows a person rotating their head. Among all methods, our method and \textbf{MC} are the only approaches capable of performing this cross-category transfer effectively. Other methods either leak the appearance from the source video or fail to transfer the motion accurately.}
  \label{fig:cross_category_1}
\end{figure}

\begin{figure}[ht]
  \centering
\includegraphics[width=\linewidth]{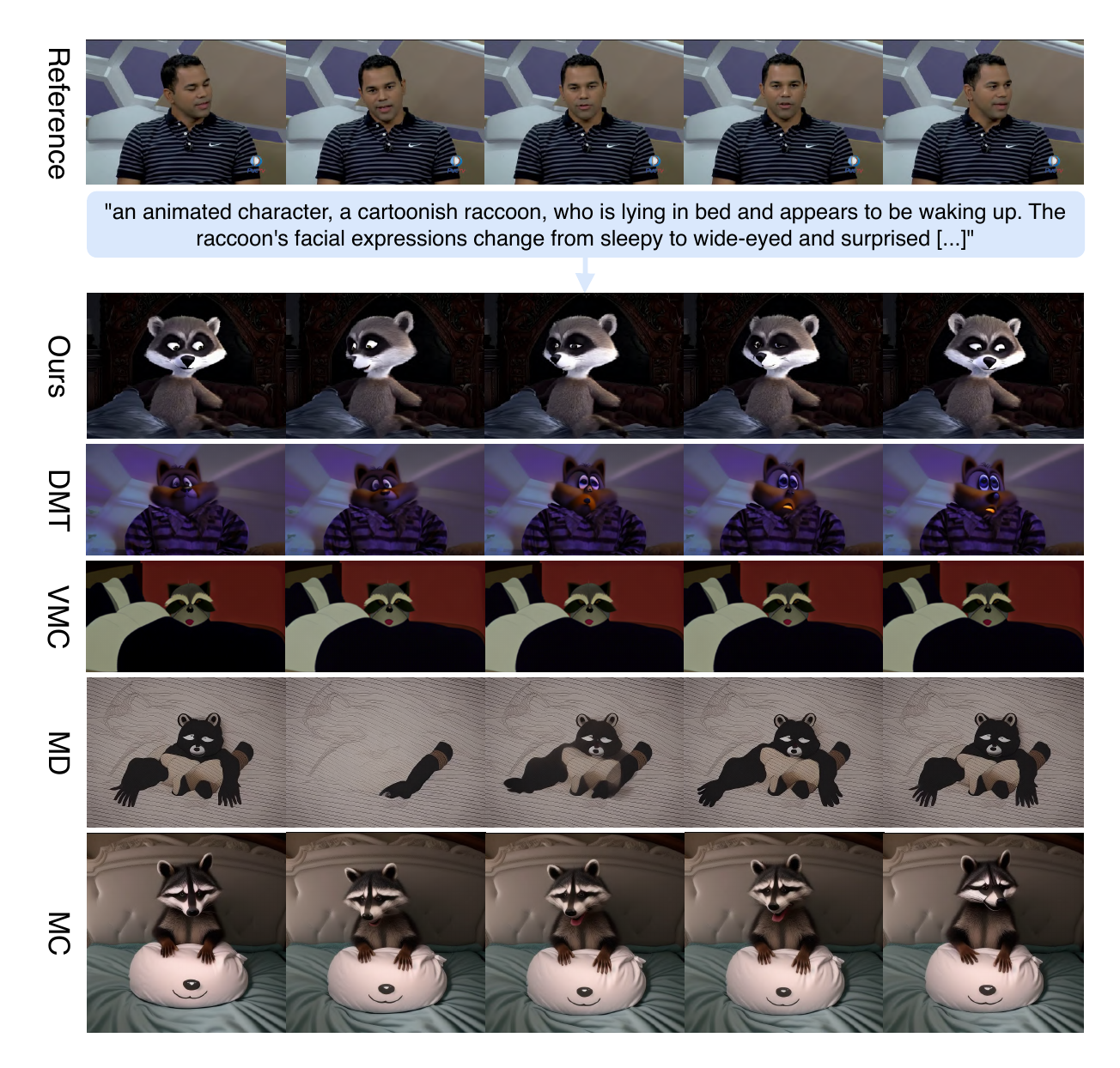}
  \caption{\textbf{Cross-category Transfer.} The reference video shows the upper body of a person performing rotating head movements. Our method successfully identifies the relative head pose changes as motion and applies them to the target character, starting from a different initial head pose. In contrast, most other methods fail to capture this motion. The only exception is \textbf{MC}, which, however, remains tied to the reference pose.}
  \label{fig:cross_category_2}
\end{figure}

\begin{figure}[ht]
  \centering
\includegraphics[width=\linewidth]{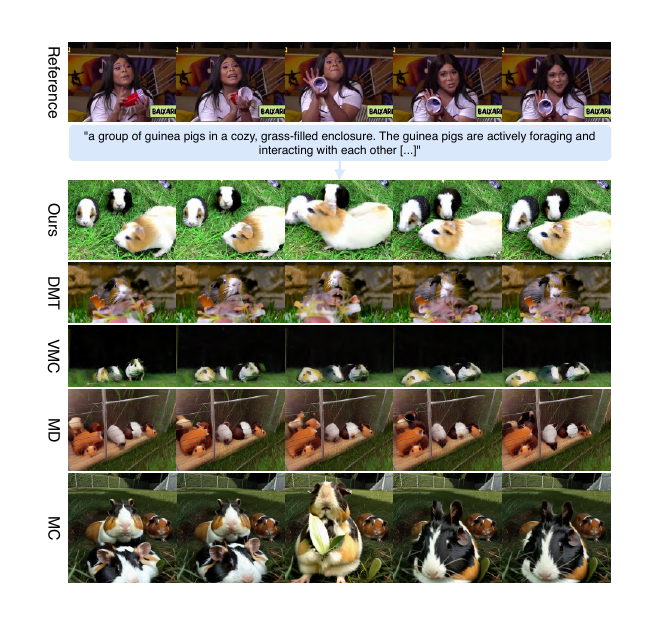}
  \caption{\textbf{Cross-category Transfer.} The source video depicts a woman holding a cup and moving it with both hands to the left side. Our method is the only approach that effectively captures this motion transfer across categories. It interprets the rotation and translation of the reference objects (cups and arms) as the movements of the guinea pigs, that serve as the general directional guidance. \textbf{MD} is the only other method that performs reasonably well, while most others fail significantly.}
  \label{fig:cross_category_3}
\end{figure}

\begin{figure}[ht]
  \centering
\includegraphics[width=\linewidth]{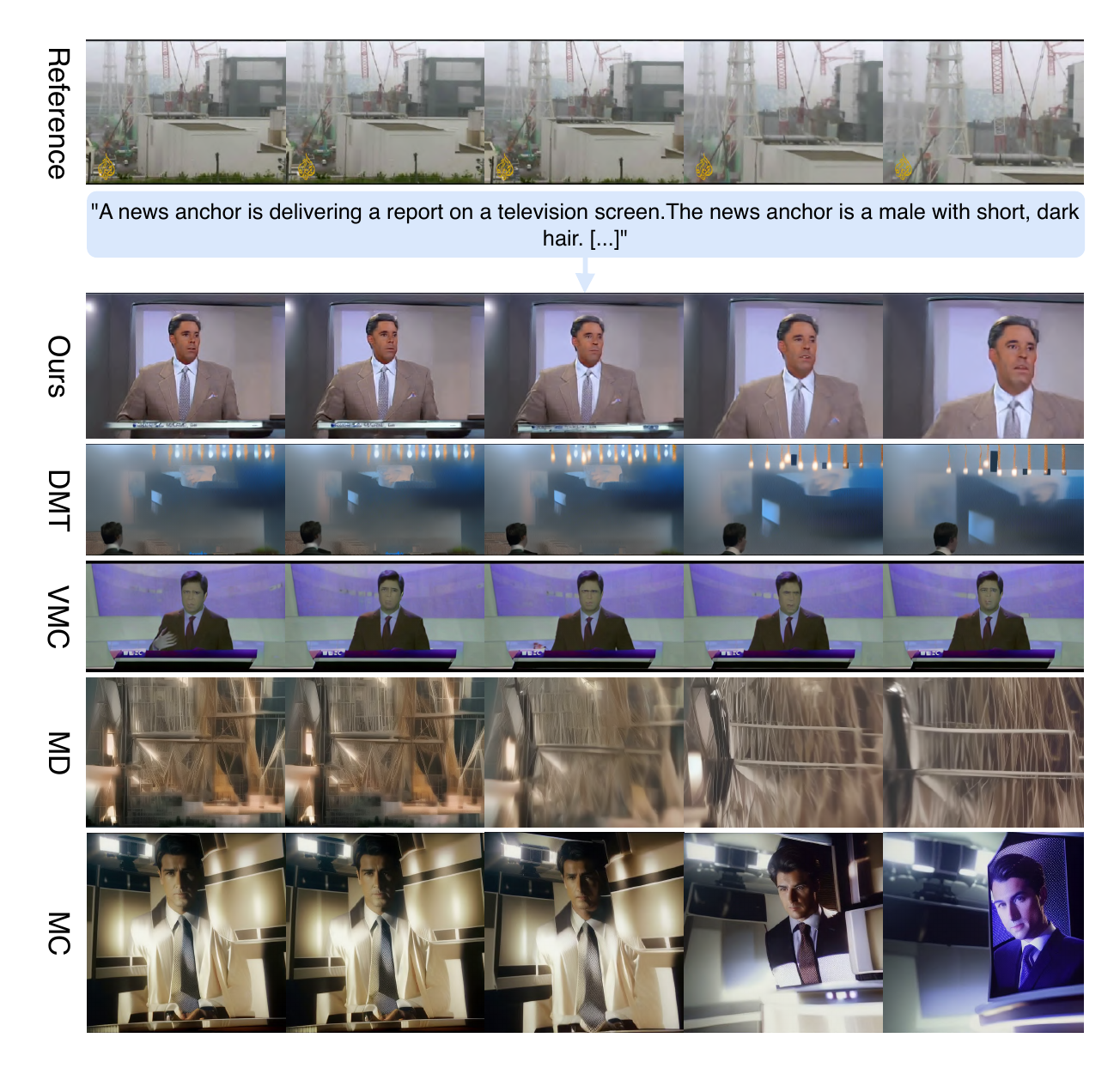}
  \caption{\textbf{Camera Motion Transfer.} The reference video solely shows a zoom-in to the top left of the frame. Our method captures this motion even while transfering to a substantially different target category, while the others fail to capture it or even fail to produce realistic videos due to the strong change in category.}
  \label{fig:camera_motion_1}
\end{figure}

\begin{figure}[ht]
  \centering
\includegraphics[width=\linewidth]{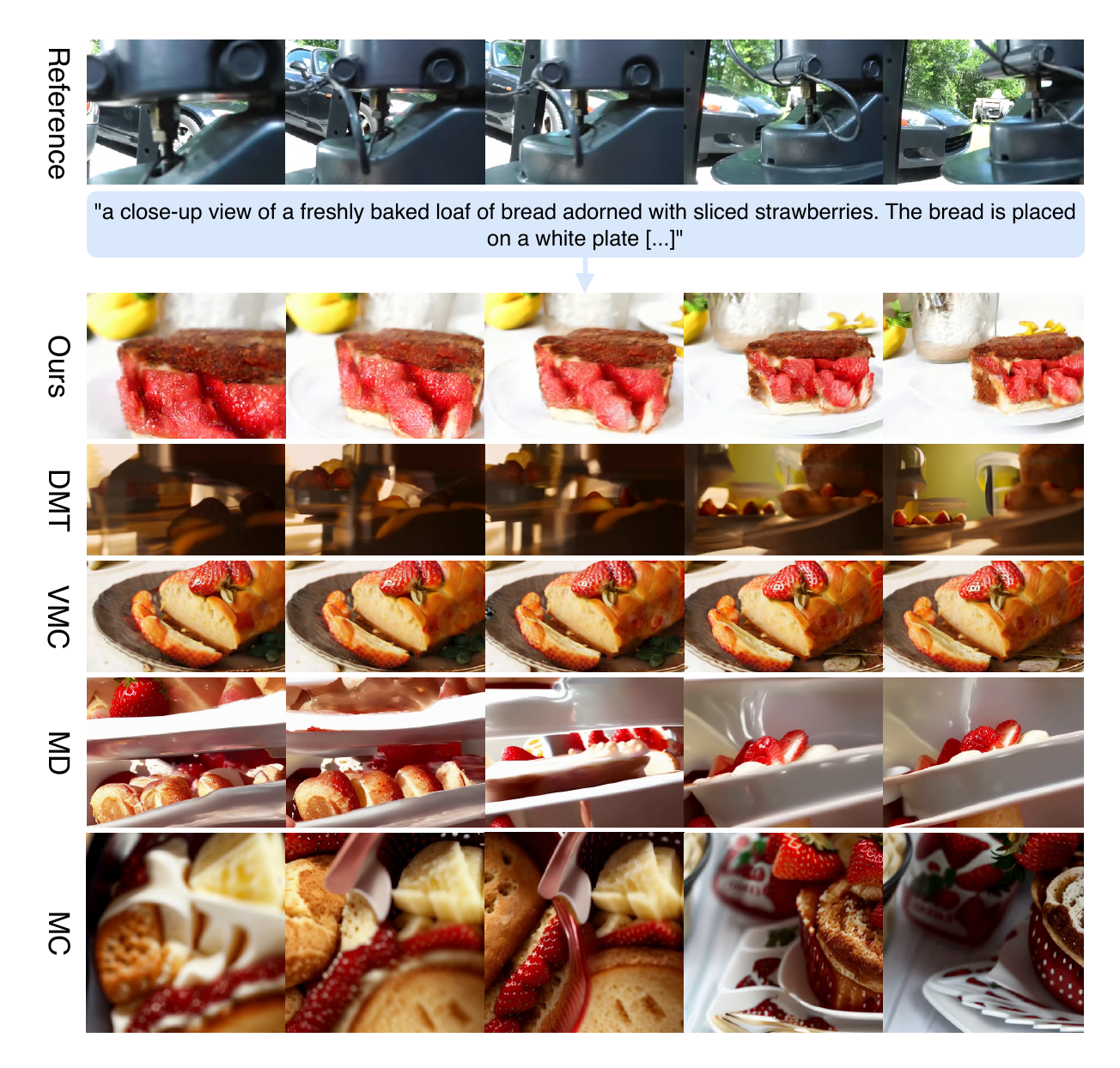}
  \caption{\textbf{Camera Motion Transfer.} The reference video depicts a camera zoom-out combined with a rotation. Our method is the only approach capable of capturing both the intended motion while maintaining a coherent and realistic generated video.}
  \label{fig:camera_motion_2}
\end{figure}

\begin{figure}[ht]
  \centering
\includegraphics[width=\linewidth]{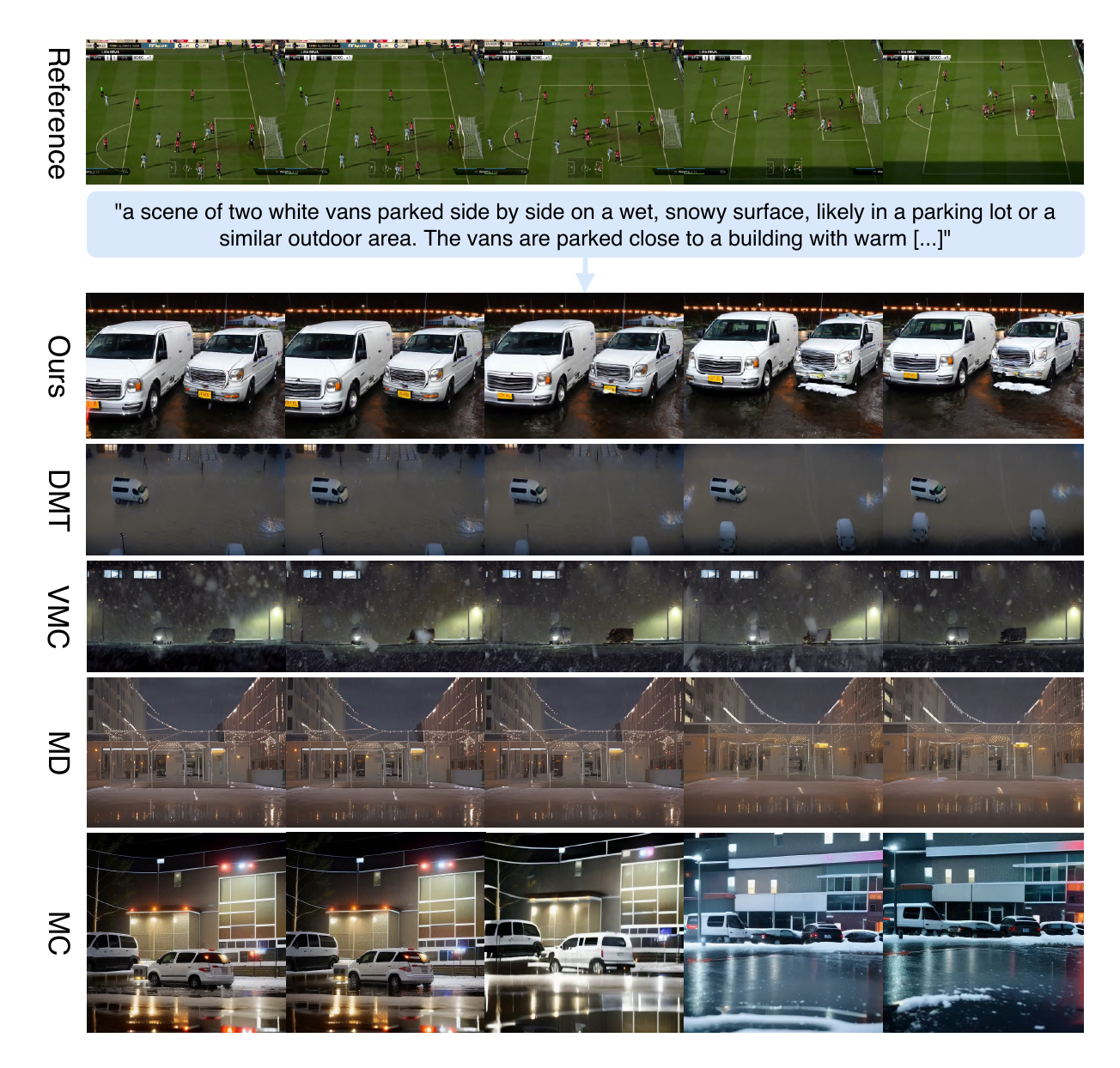}
  \caption{\textbf{Camera Motion Transfer.} The reference video depicts a soccer match with camera moving backward and then to the right. Our approach is the only method that faithfully captures this motion, while adhering to the text prompt, notably while filtering out unnecessary noise from the individual soccer players. Other methods either fail to replicate the motion accurately or do not align with the prompt.}
  \label{fig:camera_motion_3}
\end{figure}

In \cref{fig:same_category_1}, \cref{fig:same_category_2}, \cref{fig:same_category_3}, and \cref{fig:same_category_4}, we show that our approach outperforms existing methods in same-category motion transfer, especially when structural differences exist between the reference and generated videos. Additionally, our abstract motion representations enable effective cross-category transfer, allowing motion to be applied across semantically different domains. We demonstrate this capability in \cref{fig:cross_category_1}, \cref{fig:cross_category_2} and \cref{fig:cross_category_3}, where motion is successfully transferred across categories. In contrast, existing methods often struggle in such scenarios, as they typically rely heavily on the structure of the input video and face a trade-off between accurately transferring motion and remaining faithful to the content described in the text prompt. Our method is decoupled from the reference structure and instead describes motion as the change between states. This becomes apparent in \cref{fig:cross_category_2}, where DisMo applies the change of the pose, rather than aligning the structure. We showcase that our approach performs extraordinarily well for camera motion transfer while being completely independent of the reference structure. This capability is demonstrated in \cref{fig:camera_motion_1}, \cref{fig:camera_motion_2} and \cref{fig:camera_motion_3}, which highlight examples of successful camera motion transfer.

\begin{figure}[ht]
  \centering
\includegraphics[width=\linewidth]{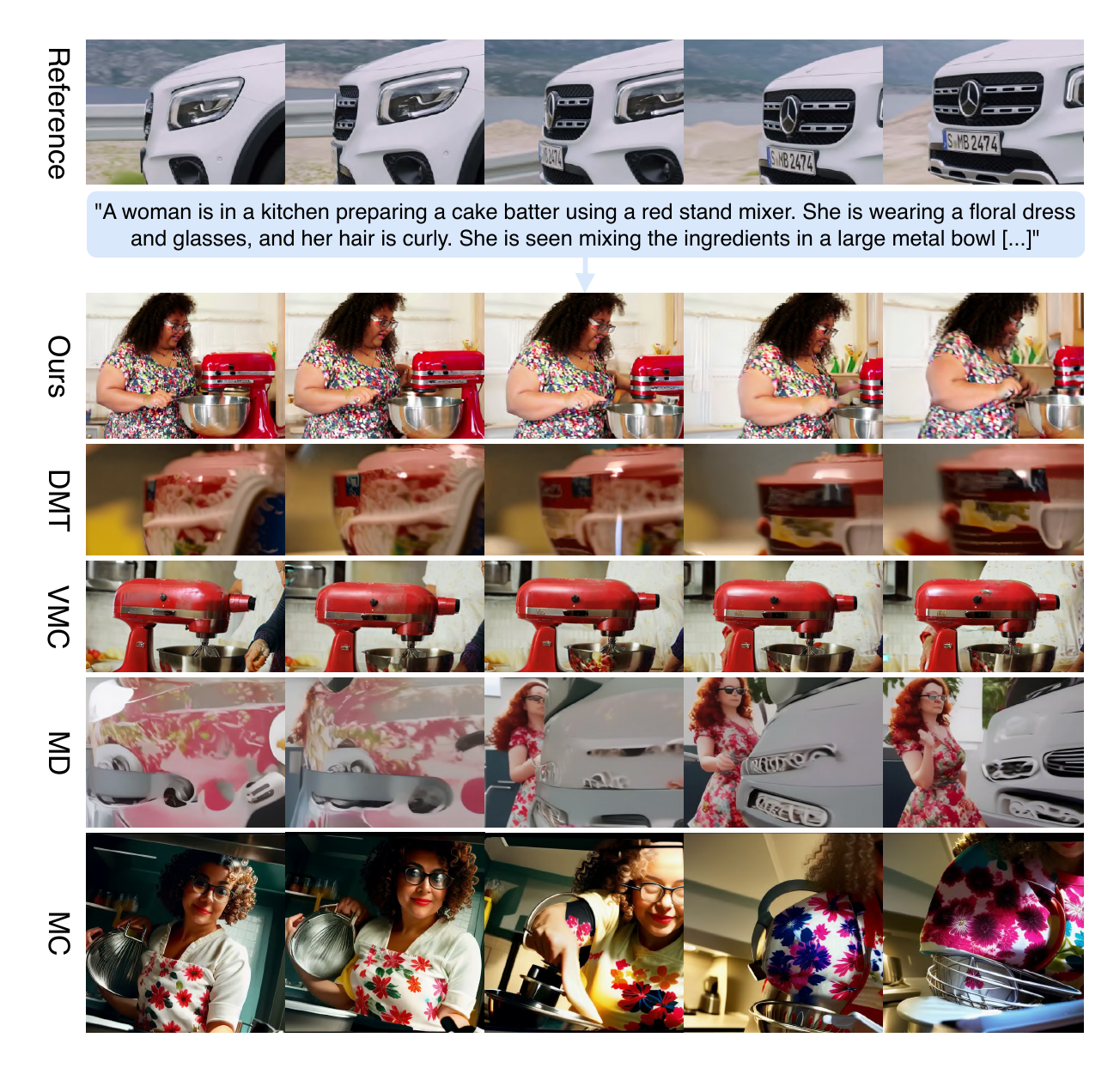}
  \caption{\textbf{Camera Motion Transfer.} The reference video depicts a rotating camera movement around a car. Our method successfully transfers this motion while remaining faithful to the text prompt. In contrast, other methods either overly rely on the input structure (e.g., \textbf{DMT}) or even leak visual appearance (e.g., \textbf{MD}), failing to reflect the prompt accurately. While \textbf{VMC} and \textbf{MC} capture aspects of the prompt, they largely disregard the reference motion.
}
  \label{fig:qual_1}
\end{figure}

\clearpage

\end{document}